\documentclass[11pt,a4paper]{article}
\usepackage[hyperref]{acl2021}
\usepackage{times}
\usepackage{latexsym}

\usepackage{microtype}

\aclfinalcopy 
\def\aclpaperid{3556} 

\usepackage{float}
\usepackage{amsmath}
\usepackage{booktabs}
\usepackage{graphicx}
\usepackage{multirow}
\usepackage{subcaption}
\usepackage{todonotes}
\usepackage{tikz}
\usetikzlibrary{matrix}
\usetikzlibrary{decorations,calligraphy}

\newcommand\rightlast{\leftskip0ptplus1fil
\rightskip0ptplus-1fil\parfillskip0ptplus1fil}
\DeclareCaptionJustification{rightlast}{\rightlast}

\makeatletter
\renewcommand{\paragraph}{%
  \@startsection{paragraph}{4}%
  {\z@}{.2ex \@plus .2ex \@minus .2ex}{-1em}%
  {\normalfont\normalsize\bfseries}%
}
\makeatother

\newcommand{\Note}[2]{} 
\newcommand{\SideNote}[2]{} 
\renewcommand{\Note}[2]{\todo[color=#1,size=\small, inline=true]{#2}} 
\renewcommand{\SideNote}[2]{\todo[color=#1,size=\small]{#2}} %

\title{Intrinsic Bias Metrics Do Not Correlate with Application Bias}

\author{Seraphina Goldfarb-Tarrant\thanks{\hspace{0.2cm}Equal contribution. Correspondence to \url{s.tarrant@ed.ac.uk}}\hspace{0.3em}$^\dagger$ \qquad
Rebecca Marchant$^*$$^\dagger$ \qquad
Ricardo Muñoz Sánchez$^*$$^\dagger$\\
\textbf{Mugdha Pandya}$^*$$^\dagger$$^\S$\qquad
\textbf{Adam Lopez}$^\ddagger$$^\dagger$\\
$^\dagger$\normalfont{University of Edinburgh, $^\ddagger$Rasa Technologies GmbH, $^\S$MACS, Heriot-Watt University}\\
   \texttt{s.tarrant@ed.ac.uk}\\
   {\tt  \{rebecca.marchant31, ricardoms.math, pandya.mugdha4\}@gmail.com}\\
   {\tt  a.lopez@rasa.com}\\
}

\date{\today}

\begin{document}
\maketitle
\begin{abstract}
Natural Language Processing (NLP) systems learn harmful societal biases that cause them to amplify inequality as they are deployed in more and more situations. To guide efforts at debiasing these systems, the NLP community relies on a variety of metrics that quantify bias in models. Some of these metrics are \textit{intrinsic}, measuring bias in word embedding spaces, and some are \textit{extrinsic}, measuring bias in downstream tasks that the word embeddings enable. Do these intrinsic and extrinsic metrics correlate with each other? We compare intrinsic and extrinsic metrics across hundreds of trained models covering different tasks and experimental conditions. Our results show \textit{no reliable correlation} between these metrics that holds in all scenarios across tasks and languages. We urge researchers working on debiasing to focus on extrinsic measures of bias, and to make using these measures more feasible via creation of new challenge sets and annotated test data. To aid this effort, we release code, a new intrinsic metric, and an annotated test set focused on gender bias in hate speech.\footnote{\url{https://tinyurl.com/serif-embed}}   
\end{abstract}

\section{Introduction}
Awareness of bias in Natural Language Processing (NLP) systems has rapidly increased as more and more systems are discovered to perpetuate societal unfairness at massive scales. This awareness has prompted a surge of research into measuring and mitigating bias, but this research suffers from lack of consistent metrics that discover and measure bias. Instead, work on bias is ``rife with unstated assumptions'' \cite{Blodgett2020LanguageI} and relies on metrics that are easy to measure rather than metrics that meaningfully detect bias in applications.

\begin{figure}[H]
  \centering
  \begin{subfigure}[t]{0.43\textwidth}
    \includegraphics[width=\textwidth]{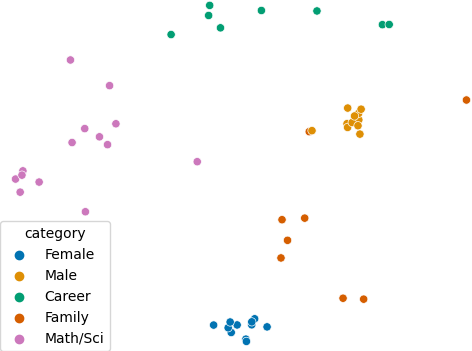}
    \caption{Intrinsic metrics summarize biases in the geometry of embeddings. For example, in this embedding space, male words are closer to words about career and about math \& science, whereas female words are closer to words about family.}
    \label{fig:intrinsic_bias}
    \end{subfigure} \hfill
    \begin{subfigure}[t]{0.45\textwidth}
    \includegraphics[width=\textwidth]{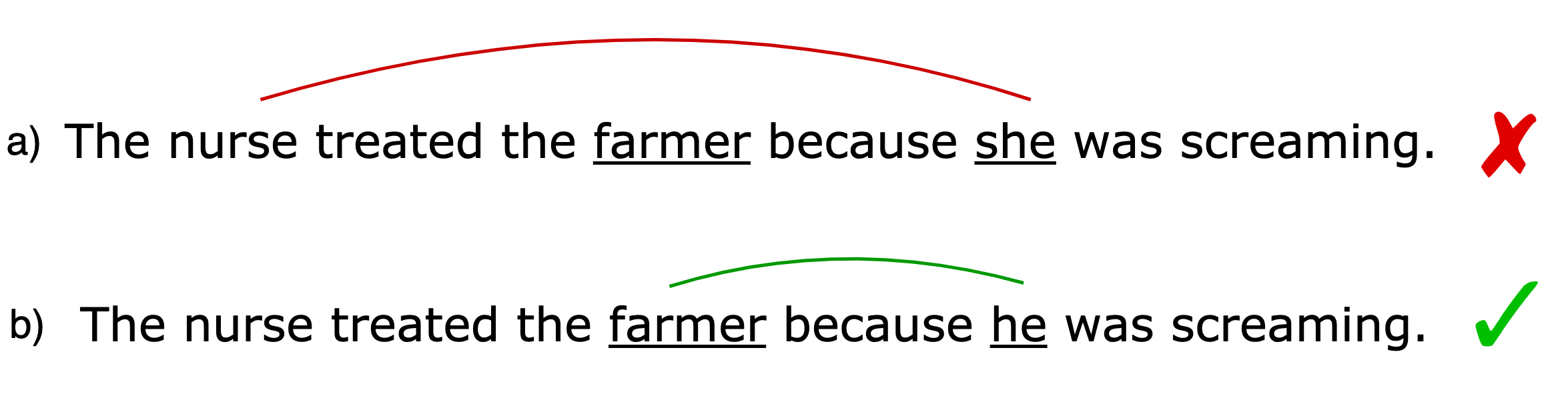}
    \caption{Extrinsic bias metrics summarize disparities in application performance across populations, such as rates of false negatives between different gender groups. For example, a coreference system may make more errors in an anti-stereotypical career coreferent (red arc) than in a pro-stereotypical one (green arc).
    }
    \label{fig:extrinsic_bias}
    \end{subfigure}
    \vspace{-.5em}
    \caption{The relationship between \textit{intrinsic} bias metrics (a) and \textit{extrinsic} bias metrics (b) has been assumed, but not confirmed.}
    \label{fig:examples}
    \vspace{-1em}
\end{figure}

A recent comprehensive survey of bias in NLP \citep{Blodgett2020LanguageI}
found that one third of all research papers focused on bias in word embeddings. This makes embeddings the most common topic in studies of bias --- over twice as common as any other topic related to bias in NLP. 
As is visualised in Figure \ref{fig:intrinsic_bias}, bias in embedding spaces is measured with \textit{intrinsic} metrics, most commonly with the Word Embedding Association Test (WEAT) \cite{Caliskan2017SemanticsDA}, which relates bias to the geometry of the embedding space. Once embeddings are  incorporated into an application,  bias can be measured via \textit{extrinsic} metrics (Figure \ref{fig:extrinsic_bias}) that 
test whether the application performs differently on language related to different populations. Hence, research on debiasing embeddings relies crucially on a hypothesis that doing so will remove or reduce bias in downstream applications. However, we are aware of \textit{no} prior research that confirms this hypothesis.

This untested assumption leaves NLP bias research in a precarious position. Research into the \emph{semantics} of word embeddings has already shown that intrinsic metrics \citep[e.g. using analogies and semantic similarity, as in ][]{Hill2015SimLex999ES} do not correlate well with extrinsic metrics \cite{Faruqui2016ProblemsWE}.
Research into the bias of word embeddings lacks the same type of systematic study, and thus as a field we are exposed to three large risks: 1) making misleading claims about the fairness of our systems, 2) concentrating our efforts on the wrong problem, and most importantly, 3) feeling a false sense of security that we are making more progress on the problem than we are. Our bias research can be rigorous and innovative, but unless we understand the limitations of metrics we use to evaluate it, it might have no impact.  




In this paper, we ask:
\textbf{Does the commonly used intrinsic metric for embeddings (WEAT) correlate with extrinsic metrics of application bias?}
To answer this question, we analyse the relationship between intrinsic and extrinsic bias. Our study considers two languages (English and Spanish), two common embedding algorithms (word2vec and fastText) and two downstream tasks (coreference resolution and hatespeech detection).

While we find a moderately high correlation between these metrics in a handful of conditions, we find no correlation or even negative correlation in most conditions. 
Therefore, we recommend that the ethical scientist or engineer does not rely on intrinsic metrics when attempting to mitigate bias, but instead focuses on the harms of specific applications and test for bias directly. 

As additional contributions to these findings, we release new WEAT metrics for Spanish, and a new gender-annotated test set for hatespeech detection for English, both of which we created in the course of this research.

\section{Bias Metrics}
\label{sec:bg}
In all of our experiments, we compute correlations between commonly-used metrics, both intrinsic and extrinsic.
\subsection{Intrinsic bias metrics}
\label{subsec:intrinsic}
Intrinsic bias metrics are applied directly to word embeddings, formulating bias in terms of geometric relationships between
\textit{concepts} such as \textit{male}, \textit{female}, \textit{career}, or \textit{family}.
Each concept is in turn represented by curated wordlists. For example, the concept \textit{male} is represented by words like \textit{brother, father, grandfather,} etc. while the concept \textit{math \& science} is represented by words like \textit{programmer, engineer,} etc.

The most commonly used metric is WEAT \cite{Caliskan2017SemanticsDA}.\footnote{We count 34 papers from *CL and FAT* conferences since January 2020 that use WEAT or SEAT \citep{May2019OnMS} in their methodology.}, which measures the difference in mean cosine similarity between two \emph{target} concepts $X$ and $Y$; and two \emph{attribute} concepts $A$  and $B$. This difference represents the imbalance in associations between concepts. 
Using $\vec{w}$ to represent the embedding of word $w$, we have a \textit{test statistic}:
\begin{align*}
    s(X,Y,A,B) = \sum_{x\in{X}}s(x,A,B) - \sum_{y\in{Y}}s(y,A,B)
\end{align*} where
\begin{align*}
    s(w,A,B) = \operatornamewithlimits{mean}_{a\in{A}}\cos(\vec{w},\vec{a}) - \operatornamewithlimits{mean}_{b\in{B}}\cos(\vec{w},\vec{b})
\end{align*}
This is normalised by the standard deviation to get the \textit{effect size} which we use in our experiments.

WEAT was initially developed as an \textit{indicator} of bias, to show that the Implicit Association Test (IAT) from the field of psychology \cite{Greenwald1998MeasuringID} can be replicated via word embeddings measurements. There are thus 10 original tests chosen to replicate the tests presented to human subjects in IAT. The tests measure different kinds of biased associations, such as African-American names vs. White names with pleasant vs. unpleasant terms, and female terms vs. male terms with career vs. family words.

WEAT was later repurposed as a \textit{predictor} of bias in embedding spaces,
via a somewhat muddy logical journey. 
It has since been translated into 6 other languages \citep[XWEAT;][]{Lauscher2019AreWC}, and extended to operate on full sentences \cite{May2019OnMS} and on contextual language models \cite{Kurita2019QuantifyingSB}. When WEAT is used as a metric, papers report the effect size of the subset of tests relevant to the task at hand, each separately.

There are known issues with WEAT, such as sensitivity to corpus word frequency, and sensitivity to target and attribute wordlists, as found by \citet{sedoc-ungar-2019-role} and \citet{ethayarajh-etal-2019-understanding}. The latter proposes an alternative more theoretically robust metric, relational inner product association (RIPA), which uses the principal component of a gender subspace (determined via the method of \citet{Bolukbasi2016ManIT}) to directly measure how "gendered" a word is. We have chosen to use the most common version of WEAT for this first empirical study, since it is most widely used. It would be interesting to test RIPA in the same way, if it were extended to more types of bias and more languages. But we note that all intrinsic metrics are sensitive to chosen wordlists, so this must be done carefully, especially across languages, a topic we will return to in Section \ref{subsec:weat_es}. 

\subsection{Extrinsic bias metrics}
\label{subsec:extrinsic}
Extrinsic bias metrics measure bias in applications, via some variant of performance disparity, or \textit{performance gap} between groups. For instance, a speech recognition system is unfair if it has higher error rates for African-American dialects \cite{Tatman2017GenderAD}, meaning that systems perform less well for those speakers. A hiring classification system is unfair if it has more false negatives for women than for men, meaning that more qualified women are accidentally rejected than are qualified men.\footnote{ \url{https://tinyurl.com/y6c6clzu}} There are two commonly used metrics to quantify this possible performance disparity: Predictive Parity \cite{Hutchinson201950YO}, which measures the difference in \emph{precision} for a privileged and non-privileged group, and Equality of Opportunity \cite{Hardt2016EqualityOO}, which measures the difference in \emph{recall} between those groups (see Appendix \ref{app:bias_metrics} for formal definitions). 

\begin{figure}[t]
  \includegraphics[width=0.45\textwidth]{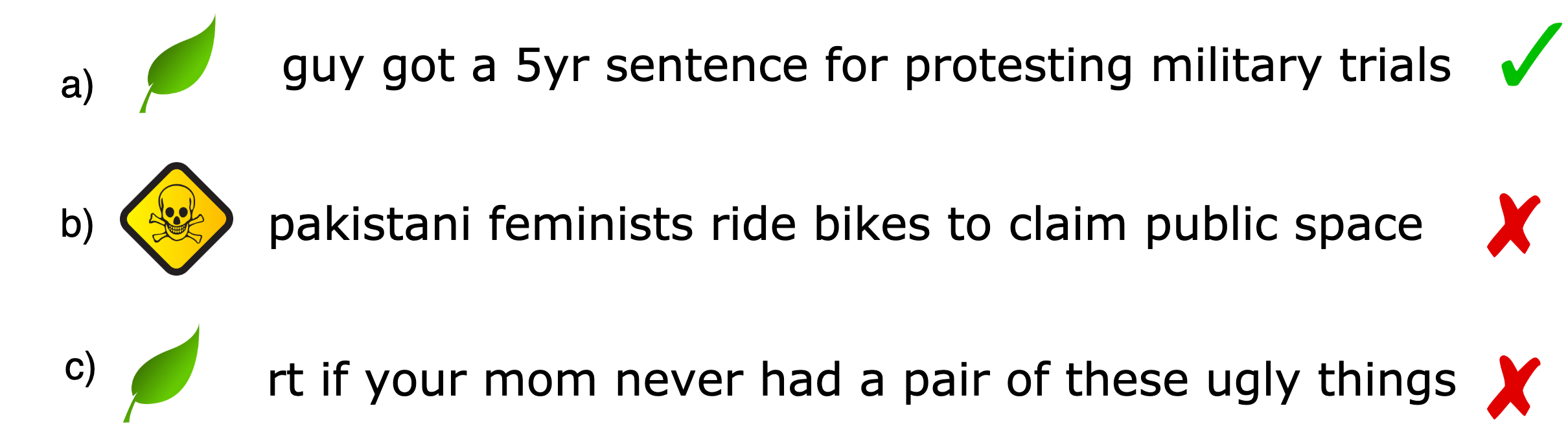}
    \vspace{-.5em}
    \caption{Examples from twitter hatespeech detection: correct (a), false positive (b), and false negative (c). This shows both kinds of problematic performance gap. b) censors harmless text and c) lets a targeted toxic comment slip through.}
    \label{fig:extrinsic_hsd}
    \vspace{-0.75em}
\end{figure}

The metric that best identifies bias in a system varies based on the task. For different applications, false negatives may be more harmful, for others false positives may be. For our first task of coreference
(Figure \ref{fig:extrinsic_bias}), false negatives --- where the system fails to identify anti-stereotypical coreference chains (e.g. women as farmers or as CEOs) --- are more harmful to the underprivileged class than false positives. For our second task, hate speech detection (Figure \ref{fig:extrinsic_hsd}), both can be harmful, for different reasons. False positives for one group can systematically censor certain content, as has been found for hate speech detection applied to African-American Vernacular English (AAVE) \cite{Sap2019TheRO, Davidson2019RacialBI}. False negatives permit abuse of minority populations that are targets of hate speech. We examine performance gaps in both precision and in recall for broad coverage.


\section{Methodology}
Each of our experiments measures the correlation between a specific instance of WEAT and a specific extrinsic bias metric. In each experiment, we train an embedding, measure the bias according to WEAT, and measure the bias in the downstream task that uses that embedding. We then modify the embeddings by applying an algorithm to either \emph{debias} them, or --- by inverting the algorithm's behavior --- to \emph{overbias} them. Again we measure WEAT on the modified embedding and also the downstream bias in the target task. When we have done this multiple times until we reach a stopping condition (detailed below), we compute the correlation between the two metrics (via Pearson correlation and analysis with scatterplots).

Rather than draw conclusions from a single experiment, we attempt to draw more robust conclusions by running many experiments, which vary along several dimensions. We  consider two common embedding algorithms, two tasks, and two languages. A full table of experiment conditions can be found in Table \ref{tab:experiments}.

\begin{table*}[t]
\centering
\resizebox{\textwidth}{!}{%
\begin{tabular}{cccc}
\toprule
\textbf{Task} & \textbf{Data} & \textbf{Bias Type} & \textbf{Intrinsic Metrics} \\ \midrule
English Coreference & Ontonotes/WinoBias & Gender & WEAT 6, 7, 8 \\  
English Hate speech & Twitter & Gender & WEAT 6, 7, 8  \\  
Spanish Hate speech & Twitter & Gender &  XWEAT 7+8 (new)  \\ 
Spanish Hate speech & Twitter  &Migrants &  XWEAT Migrants (new)   \\ \bottomrule
\end{tabular}%
}
\caption{Tasks used in our experiments. Each experiment consists of a task, an embedding method (either word2vec or fasttext), an intrinsic metric (one experiment for each listed), and an extrinsic metric (either Predictive Parity or Equality of Opportunity). We run an experiment for all possible combinations. To produce data points for each experiment, we use preprocessing and post-processing methods to debias and overbias the input word embeddings.}
\label{tab:experiments}
\end{table*}

\subsection{Debiasing and Overbiasing}
We need to measure the relationship between intrinsic and extrinsic metrics as bias changes, we must generate many datapoints for each experiment. Previous work on bias in embeddings studies methods to \textit{reduce} embedding bias. To generate enough data points, we take the novel approach of both decreasing \textit{and} increasing bias in the embeddings. We measure the baseline bias level, via WEAT, for each embedding trained normally on the original corpus. We then adjust the bias up or down, remeasure WEAT, and measure the change in the downstream task.  

We choose two methods from previous work that are capable of both debiasing and overbiasing: the first is a preprocessing method that  operates on the training data before training, the second is a post-processing method that operates on the embedding space once it has been trained. This is important since both kinds of methods may be used in practice: a large company with proprietary data will train embeddings from scratch, and thus may use a preprocessing method; whereas a small company may rely on publicly available pretrained embeddings, and thus use a post-processing method.
\footnote{There are additional embedding based debiasing methods used in practice, based on identifying and removing a gender subspace during training or as postprocessing \cite{Bolukbasi2016ManIT, Zhao2018LearningGW}. However, these methods do not change a word's nearest neighbour clusters \cite{Gonen2019LipstickOA}, and so we would expect these debiasing methods to show superficial bias changes in WEAT without changing downstream bias. Both methods that we select modify the underlying word distribution and move many words in relation to each other. We verified this with tSNE visualisation as in Figure \ref{fig:intrinsic_bias} following \citet{Gonen2019LipstickOA} and find that our bias modification methods do change word clusters.}    

For preprocessing, we use dataset balancing \cite{Dixon2018MeasuringAM}, which consists of sub-sampling the training data to be more equal with respect to some attributes.
For instance, if we are adjusting gender bias, we identify pro-stereotypical sentences\footnote{Stereotypes as defined by \newcite{Zhao2018GenderBI} and by \newcite{Caliskan2017SemanticsDA}, who use the U.S. Bureau of Labor Statistics and the Implicit Association Test, respectively.}
such as `\textit{She} was a talented housekeeper' vs. anti-stereotypical sentences, such as `\textit{He} was a talented housekeeper' or `She was a talented analyst'. We sub-sample and reduce the frequency of the pro-stereotypical collocations to debias, and sub-sample the anti-stereotypical conditions to overbias.

As a postprocessing method for already trained embeddings, we use the Attract-Repel \cite{Mrksic2017SemanticSO} algorithm. This algorithm was developed to use dictionary wordlists (synonyms, antonyms) to refine semantic spaces.
It aims to move similar words (synonyms) close to each other and dissimilar words (antonyms) farther from each other, while keeping a regularisation term to preserve original semantics as much as possible.  \newcite{Lauscher2020AGF} used an approach inspired by Attract-Repel for debiasing, though with constraints implemented somewhat differently.
 We use the same pro- and anti-stereotypical wordlists as in dataset-balancing. For debiasing, we use the algorithm to increase distance between pro-stereotypical combinations (\textit{she, housekeeper}) and decrease distance between anti-stereotypical combinations (\textit{she, analyst} or \textit{he, housekeeper}). For overbiasing we do the reverse.\footnote{Wordlists used for bias-modification and configs for Attract-Repel are included in the codebase.}
 
As the stopping condition for preprocessing, we constrain the sub-sampling so that it does not substantially change the dataset size, by limiting it to removing less than five percent of the original data. For postprocessing we limit the algorithm to maximum 5 iterations.

\subsection{Embedding Algorithms} We use two common word embedding algorithms: fastText \cite{bojanowski2017enriching} and Skip-gram word2vec \cite{Mikolov2013EfficientEO} embeddings. Word embeddings in fastText are composed from embeddings of both the word and its subwords in the form of character $n$-grams. \newcite{Lauscher2019AreWC} suggest that this difference  may cause bias to be acquired and encoded differently in fastText and word2vec (We discuss this in more detail in Section \ref{sec:results}).

Despite recent widespread interest in contextual embeddings \citep[e.g. BERT;][]{devlin-etal-2019-bert}, our experiments use these simpler contextless embeddings because they are widely available in many toolkits and used in many real-world applications. Their design simplifies our experiments, whereas contextual embeddings would add significant complexity. However, we know that bias is still a problem for large contextual embeddings \cite{Zhao2019GenderBI, Zhao2020GenderBI, Gehman2020RealToxicityPromptsEN,Sheng2019TheWW}, so our work remains important. If intrinsic and extrinsic measures do not correlate with simple embeddings, this result is unlikely to be changed by adding more architectural layers and configurable hyperparameters. 

\subsection{Downstream tasks} We use three tasks that appear often in bias literature: Coreference resolution for English, hate speech detection for English, and hate speech detection for Spanish. To make the scenarios as realistic as possible, we use a common, easy-to-implement and high performing architecture for each task: the end-to-end coreference system of \newcite{Lee2017EndtoendNC} and the the CNN of \newcite{Kim2014ConvolutionalNN}, which has been used in high-scoring systems in recent hate speech detection shared tasks \cite{Basile2019SemEval2019T5}. 
For each task, we feed pretrained embeddings to the model, frozen, and then train the model using the standard hyperparameters published for each model and task.

\subsection{Languages} We experiment on both English and Spanish. It is important to take a language with pervasive gender-marking (Spanish) into account, as previous work has shown that grammatical gender-marking has a strong effect on gender bias in embeddings \cite{McCurdy2017GrammaticalGA,Gonen2019HowDG, Zhou2019ExaminingGB}. We use Spanish only for hate speech detection, because gender marking makes a challenge-set style coreference evaluation trivial to resolve and not a candidate for detection of gender bias.\footnote{This fact is the premise behind the work of \newcite{Stanovsky2019EvaluatingGB} who use the explicit marking in translation to reveal bias.} 


\section{Experiments}
\subsection{Datasets} 
\label{subsec:datasets}
To train embeddings, we use domain-matched data for each downstream task. For coreference we train on Wikipedia data, and for hatespeech detection we train on English tweets or Spanish tweets, consistent with the task.\footnote{Details of datasets \& preprocessing are in Appendix \ref{app:train_data}.} Our English Coreference system is trained on OntoNotes \cite{Weischedel2017OntoNotesA} and evaluated on Winobias \cite{Zhao2018GenderBI}, a Winograd-schema style challenge set designed to measure gender bias in coreference resolution. English hate speech detection uses the abusive tweets dataset of \newcite{Founta2018LargeSC}, and is evaluated on the test set of ten thousand tweets, which we have hand labelled as targeted \textit{male}, \textit{female}, and \textit{neutral} (we release this labelled test set for future work). Spanish hate speech detection uses the data from the shared task of \newcite{Basile2019SemEval2019T5}, which contains labels for comments directed at women and directed at migrants.

\subsection{WEAT \& Bias modification wordlists} 
\label{subsec:bias_mod_wordlists}
Both WEAT and bias modification methods depend on seed wordlists.\footnote{WEAT uses wordlists to measure relationships between words in the space, and bias modification depends on identifying words to sub or supersample (for databalancing), or to adjust (for Attract-Repel). Many other debiasing methods that we did not use (e.g \citet{Bolukbasi2016ManIT}) also use wordlists.} 
These wordlists are closely related to each other, and we match them by type of bias, such that we measure WEAT tests for gender bias with embeddings modified via gender bias wordlists (themselves derived from WEAT lists, as detailed below) and WEAT tests for migrant bias with embeddings modified for migrant bias.  

WEAT wordlists are standardised, and for English we use the three WEAT test wordlists (numbers 6,7,8) for gender.\footnote{All WEAT wordlists are in Appendix \ref{app:weat}. We make a small substitution of general gender words instead of proper names in WEAT 6, as proper names by design do not appear in our coreference task.}

To generate bias modification wordlists we follow the approach of \newcite{Lauscher2020AGF} and use a pretrained set of embeddings (from \url{spacy.io}) to expand the set of WEAT words to their 100 unique nearest neighbours. For all experiments, we take the union of all WEAT terms, expand them, and use this expanded set for both dataset balancing and for Attract-Repel.\footnote{Final word sets are 200-400 words, due to significant overlap in nearest neighbors \& manual removal of odd terms.} For gender bias in coreference and hate speech, we use terms that are male vs. female and are career, math, science, vs. family, art. For gender bias and migrant bias in Spanish hate speech, we compare male/female identity or migrant/non-migrant identity with pleasant-unpleasant term expansions.\footnote{We did additionally experiment with using the \textit{exact} WEAT terms for debiasing, and found the trends to be similar but of smaller magnitude, so we settled on expanded lists as a more realistic scenario.} 

\subsection{New Spanish WEAT}
\label{subsec:weat_es}
We substantially modified Spanish WEAT (aka XWEAT for non-English WEATs) and added entirely new terms. The reason for this is that the original XWEAT was translated from English very literally, which causes two  problems. 

The first problem with XWEAT is that many of the terms do not make sense in a Spanish speaking community --- names included in the original, like \textit{Amy}, are names in Spanish and thus were untranslated, but are uncommon and have upper class connotations not intended in the original test. Another example is \textit{firearms} translated as \textit{arma de fuego}, which while technically a correct literal translation, is not commonly used to describe weapons.\footnote{The standard would be \textit{armas}. \textit{arma de fuego} is also composed of three words, and so will not appear in any vocabulary.}

The second problem with XWEAT is that nouns on the wordlists for both abstract math and science concepts as well as abstract art concepts are almost entirely grammatically female. For instance, \textit{ciencia} (science), \textit{geometría} (geometry) are grammatically female, as are \textit{escultura} (sculpture) and \textit{novela} (novel). It is well established that for languages with grammatical gender, words that share a grammatical gender have embeddings that are closer together than words that do not \cite{Gonen2019HowDG, McCurdy2017GrammaticalGA}. So, when WEAT in English was translated into XWEAT in Spanish \cite{Glavas2019HowT}, the terms were imbalanced with regard to grammatical gender, which makes the results misleading. We balance the lists, often replacing  abstract nouns with corresponding adjectives which can take male or female form, e.g. 
\textit{científico} and \textit{científica} (scientific, male and female), such that we can use both versions to account for the effect of grammatical gender. 

Finally, we needed a metric to examine bias against migrants. Metrics for intrinsic bias must be targeted to the type of harm expected in the downstream application, and there is not an out-of-the-box WEAT test for this. So we create a new WEAT test for bias against migrants in Spanish. Following the setup of tests for racial bias in original WEAT --- based on American racial biases in English ---  we have lists of names associated with migrants vs. non-migrants, and compare them with lists of pleasant and unpleasant terms. The names are based on work of \newcite{SALAMANCA2013}, who studied ranking names as lower vs. upper class; class status is closely correlated with whether a person is a migrant. We select a subset of names in which the majority in the study agree on the class. Pleasant and unpleasant terms exist in WEAT and XWEAT, but we again modify them to balance grammatical gender.  



\section{Results}
\label{sec:results}

\newcommand{\ignore}[1]{}
\ignore{
\begin{figure*}[t!]
    \centering
    \begin{subfigure}[b]{\textwidth}
     \begin{subfigure}[c]{0.07\textwidth}
     \centering
         \caption*{WEAT\\ 6}
         \vspace{25mm}
     \end{subfigure}
    \begin{subfigure}[b]{0.225\textwidth}
        \caption*{fastText / Precision}
        \includegraphics[width=\textwidth]{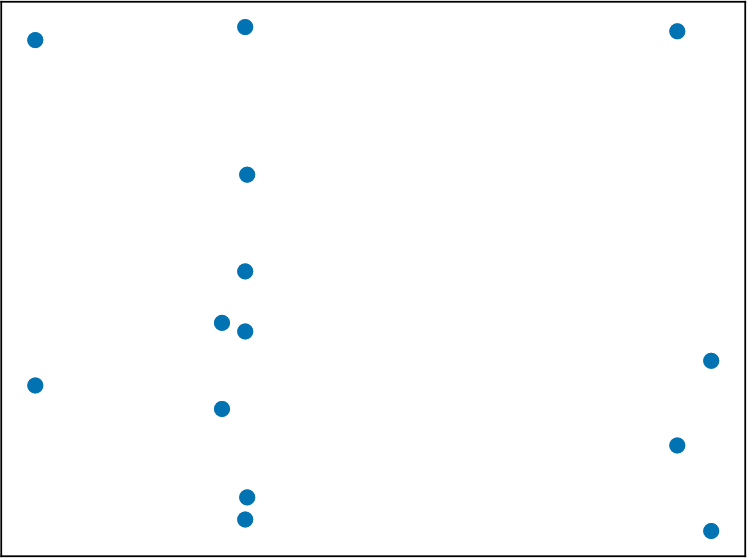}
    \end{subfigure}
    \begin{subfigure}[b]{0.225\textwidth}
    \caption*{word2vec - Precision}
        \includegraphics[width=\textwidth]{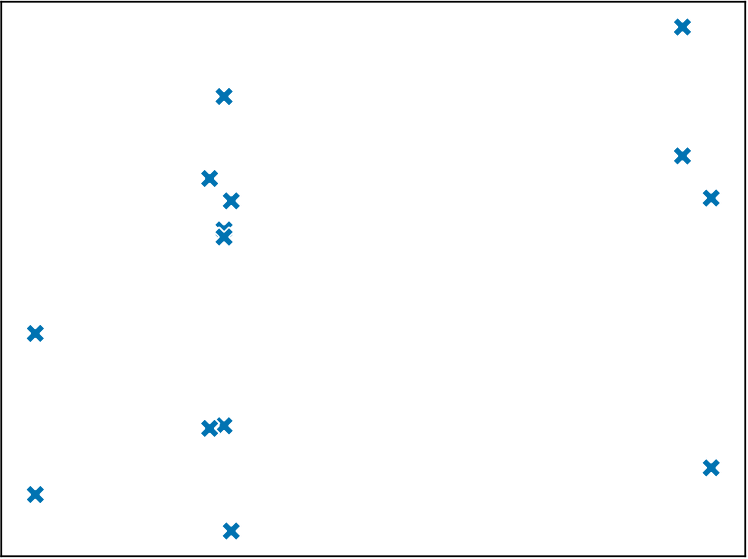}
    \end{subfigure}
    \begin{subfigure}[b]{0.225\textwidth}
    \caption*{fastText - Recall}
        \includegraphics[width=\textwidth]{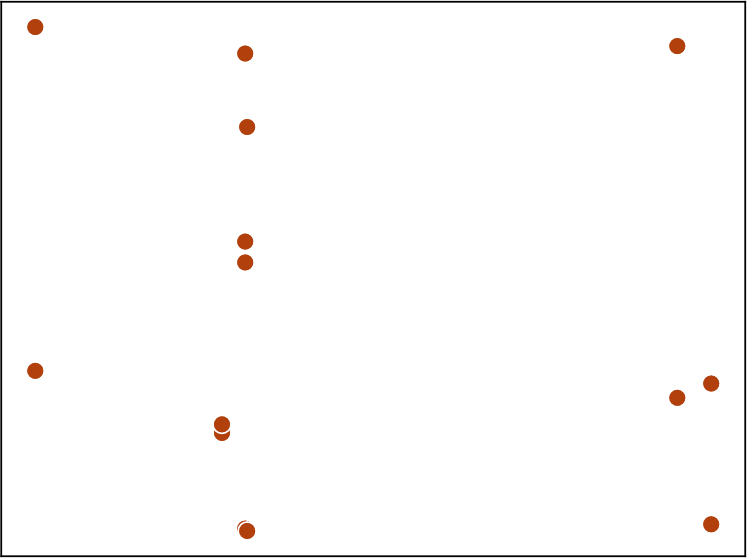}
    \end{subfigure}
    \begin{subfigure}[b]{0.225\textwidth}
    \caption*{word2vec - Recall}
        \includegraphics[width=\textwidth]{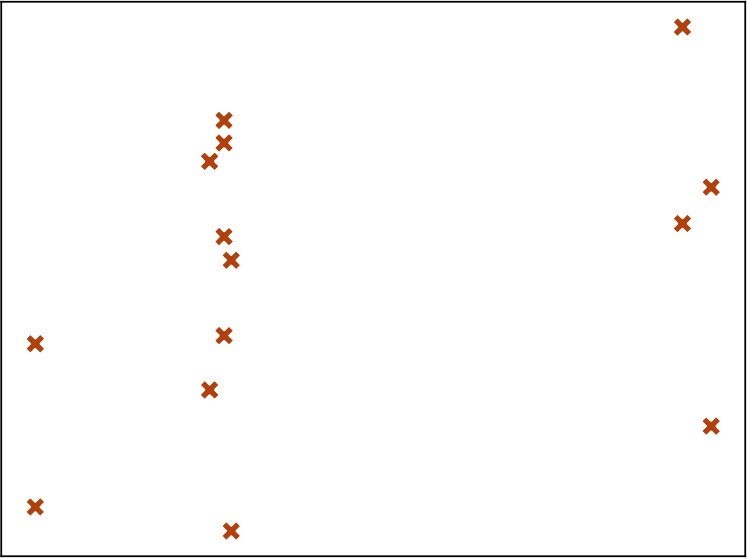}
    \end{subfigure}
    \vspace{-4em}
    \end{subfigure}
    \begin{subfigure}[b]{\textwidth}
     \begin{subfigure}[c]{0.07\textwidth}
     \centering
         \caption*{WEAT \\7}
         \vspace{25mm}
     \end{subfigure}
    \begin{subfigure}[b]{0.225\textwidth}
        \includegraphics[width=\textwidth]{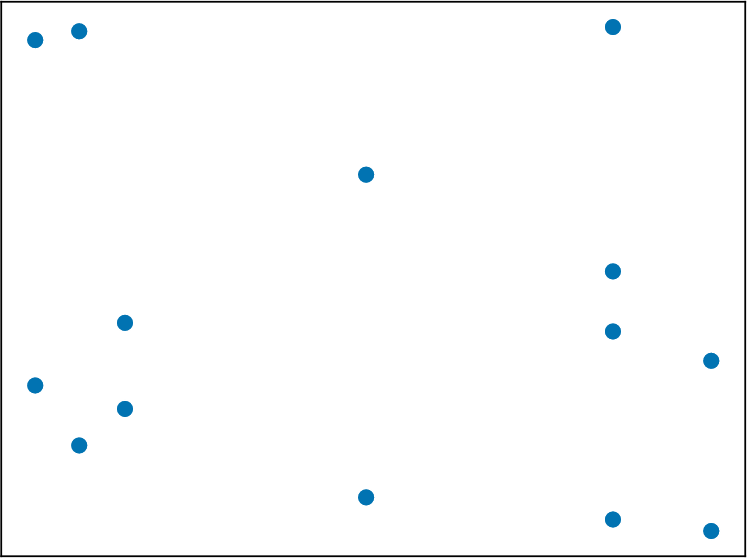}
    \end{subfigure}
    \begin{subfigure}[b]{0.225\textwidth}
        \includegraphics[width=\textwidth]{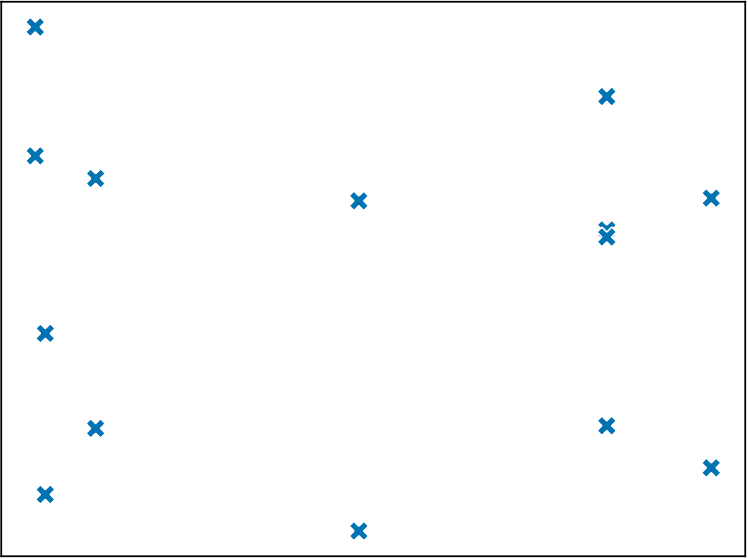}
    \end{subfigure}
    \begin{subfigure}[b]{0.225\textwidth}
        \includegraphics[width=\textwidth]{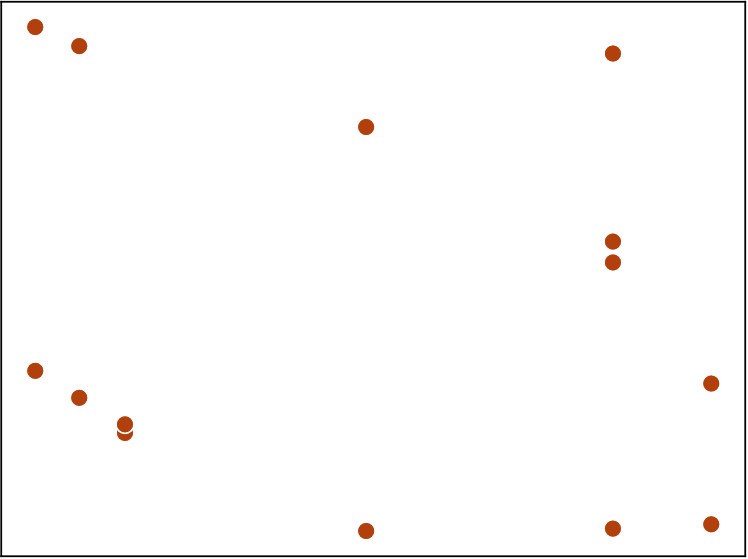}
    \end{subfigure}
    \begin{subfigure}[b]{0.225\textwidth}
        \includegraphics[width=\textwidth]{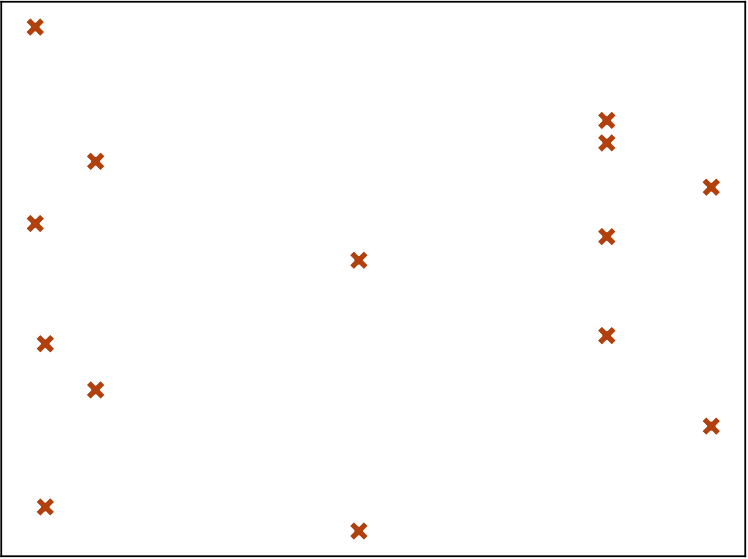}
    \end{subfigure}
    \vspace{-4em}
    \end{subfigure}
    \begin{subfigure}[b]{\textwidth}
     \begin{subfigure}[c]{0.07\textwidth}
     \centering
         \caption*{WEAT\\8}
         \vspace{25mm}
     \end{subfigure}
    \begin{subfigure}[b]{0.225\textwidth}
        \includegraphics[width=\textwidth]{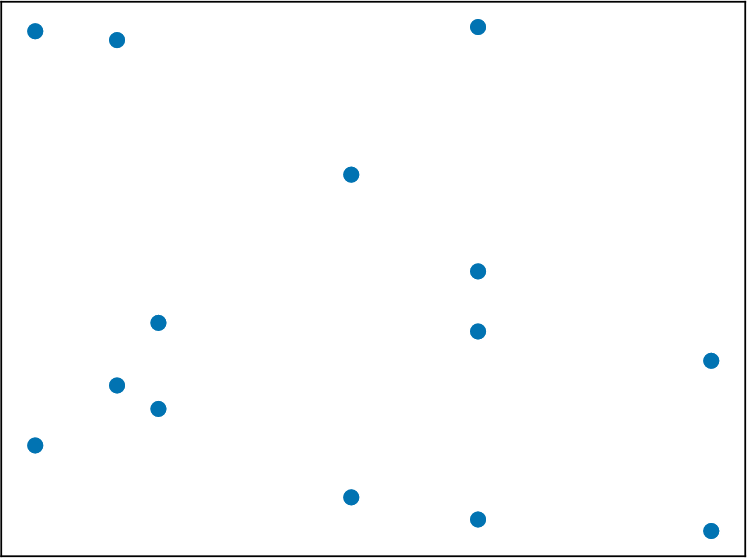}
    \end{subfigure}
    \begin{subfigure}[b]{0.225\textwidth}
        \includegraphics[width=\textwidth]{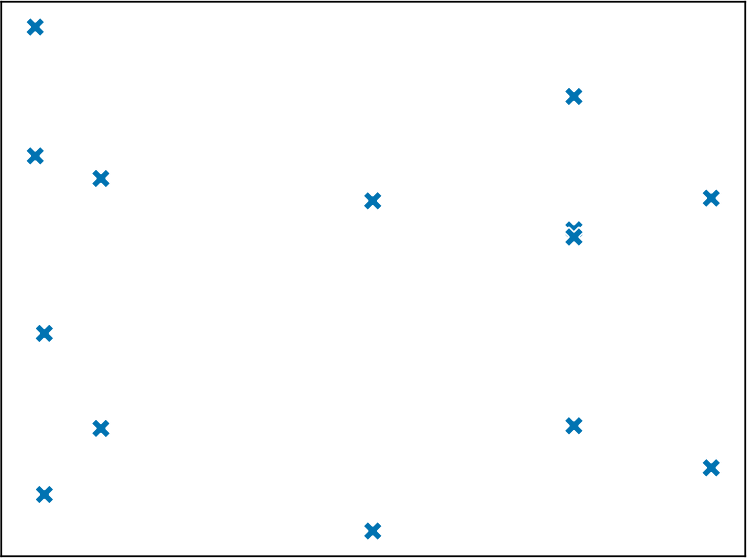}
    \end{subfigure}
    \begin{subfigure}[b]{0.225\textwidth}
        \includegraphics[width=\textwidth]{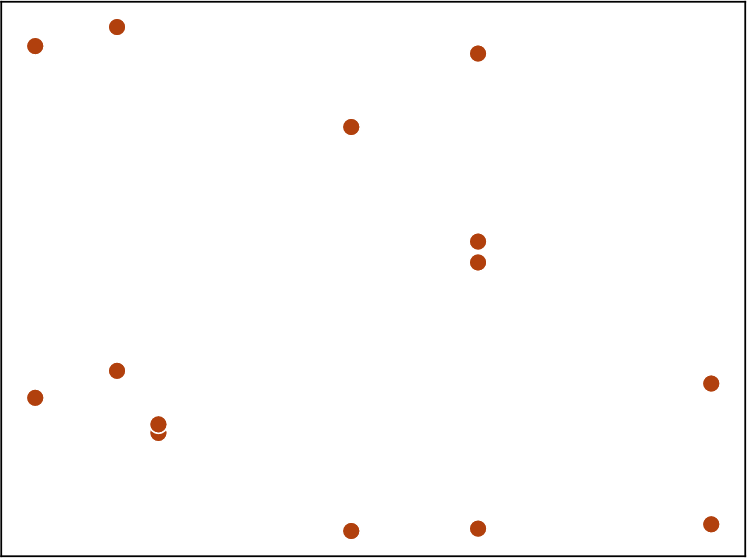}
    \end{subfigure}
    \begin{subfigure}[b]{0.225\textwidth}
        \includegraphics[width=\textwidth]{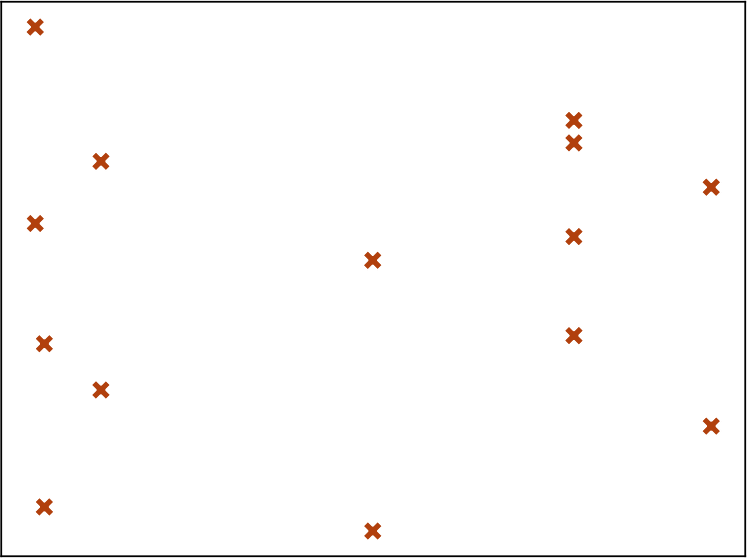}
    \end{subfigure}
    \vspace{-4.5em}
    \vspace{0.5em}
    \end{subfigure}
    \begin{subfigure}[b]{\textwidth}
     \begin{subfigure}[c]{0.07\textwidth}
     \centering
         \caption*{WEAT\\6}
         \vspace{25mm}
     \end{subfigure}
    \begin{subfigure}[b]{0.225\textwidth}
        \includegraphics[width=\textwidth]{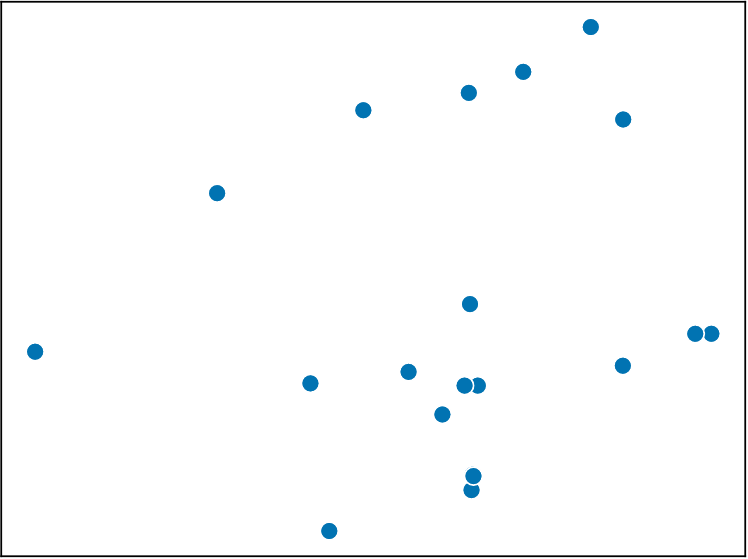}
    \end{subfigure}
    \begin{subfigure}[b]{0.225\textwidth}
        \includegraphics[width=\textwidth]{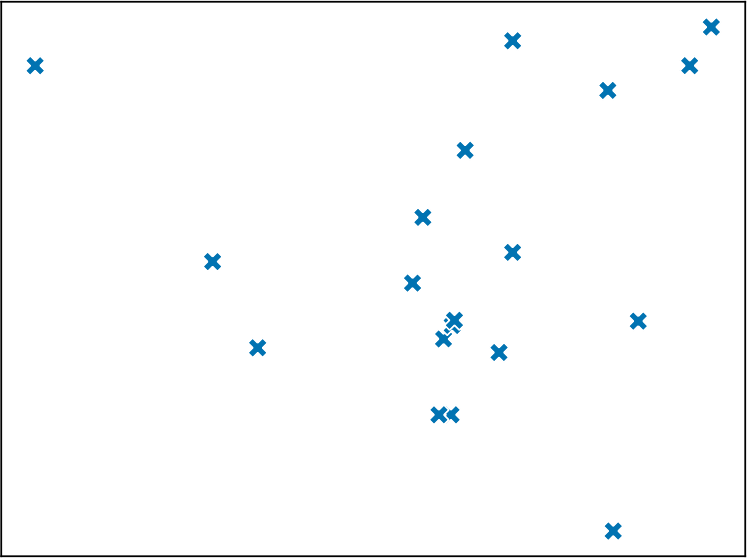}
    \end{subfigure}
    \begin{subfigure}[b]{0.225\textwidth}
        \includegraphics[width=\textwidth]{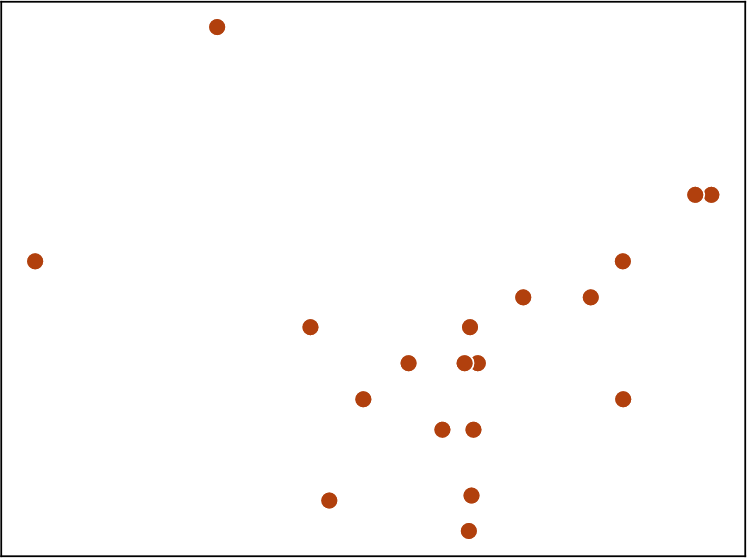}
    \end{subfigure}
    \begin{subfigure}[b]{0.225\textwidth}
        \includegraphics[width=\textwidth]{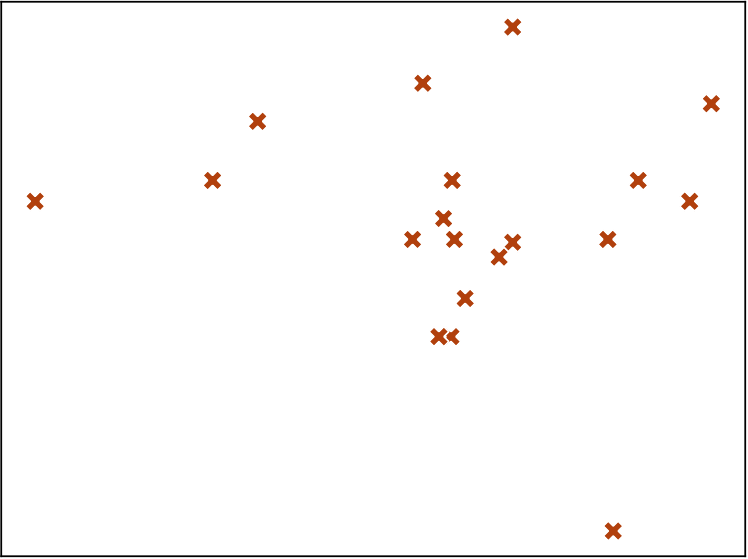}
    \end{subfigure}
    \vspace{-4em}
    \end{subfigure}
    \begin{subfigure}[b]{\textwidth}
     \begin{subfigure}[c]{0.07\textwidth}
     \centering
         \caption*{WEAT\\7}
         \vspace{25mm}
     \end{subfigure}
    \begin{subfigure}[b]{0.225\textwidth}
        \includegraphics[width=\textwidth]{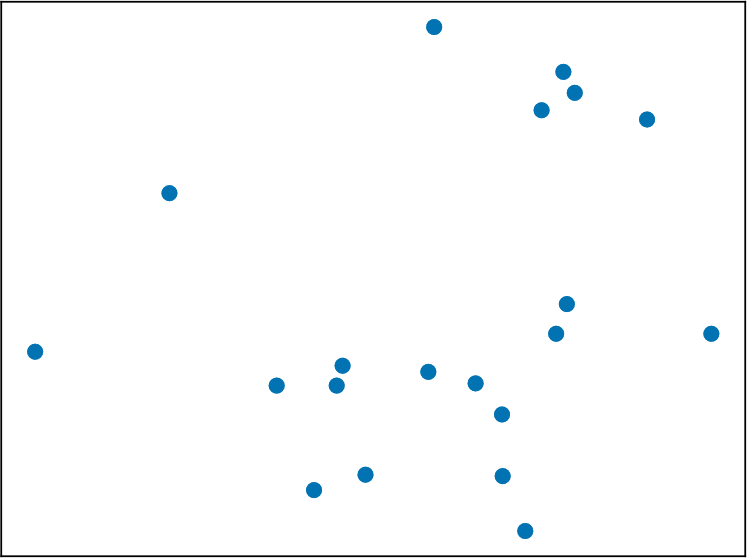}
    \end{subfigure}
    \begin{subfigure}[b]{0.225\textwidth}
        \includegraphics[width=\textwidth]{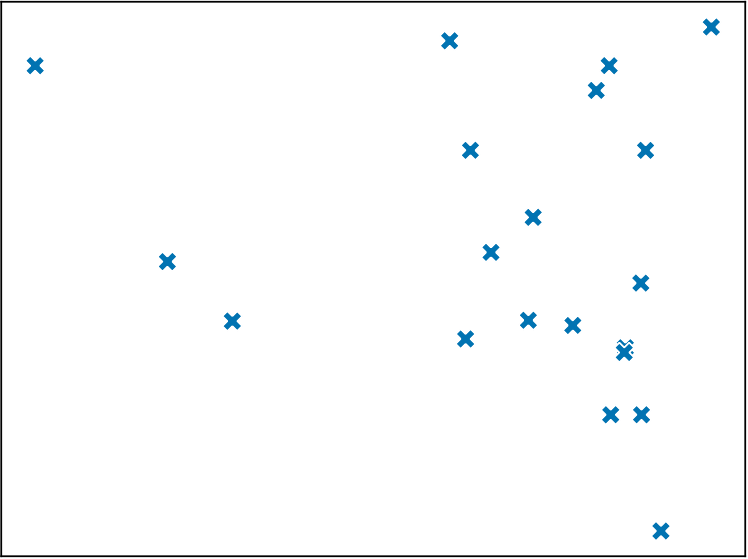}
    \end{subfigure}
    \begin{subfigure}[b]{0.225\textwidth}
        \includegraphics[width=\textwidth]{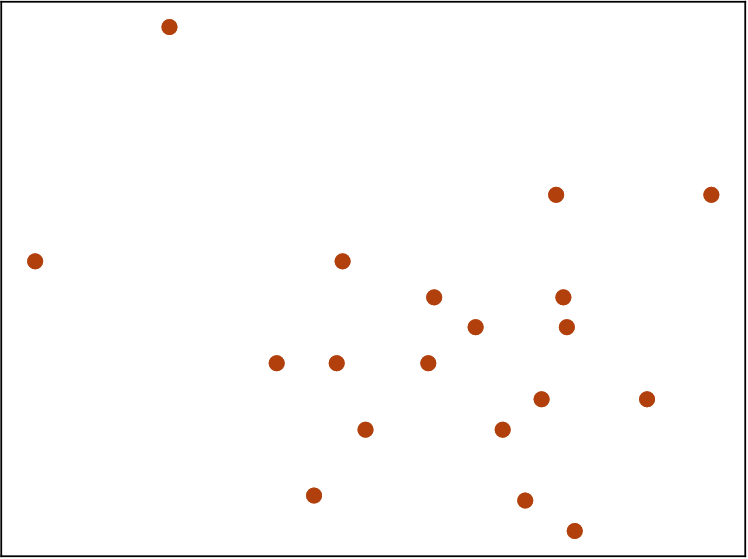}
    \end{subfigure}
    \begin{subfigure}[b]{0.225\textwidth}
        \includegraphics[width=\textwidth]{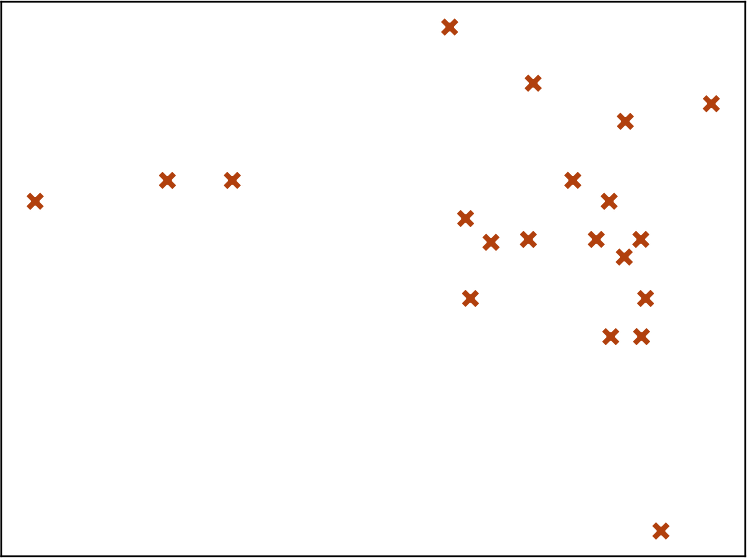}
    \end{subfigure}
    \vspace{-4em}
    \end{subfigure}
    \begin{subfigure}[b]{\textwidth}
     \begin{subfigure}[c]{0.07\textwidth}
     \centering
         \caption*{WEAT\\8}
         \vspace{25mm}
     \end{subfigure}
    \begin{subfigure}[b]{0.225\textwidth}
        \includegraphics[width=\textwidth]{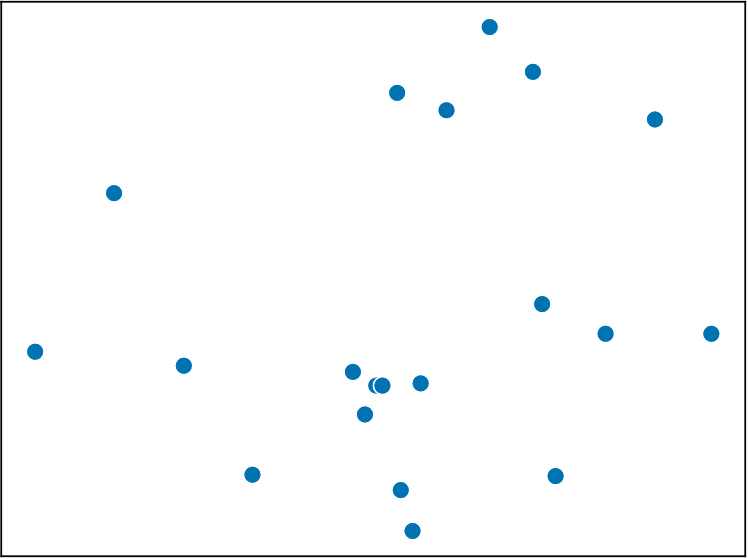}
    \end{subfigure}
    \begin{subfigure}[b]{0.225\textwidth}
        \includegraphics[width=\textwidth]{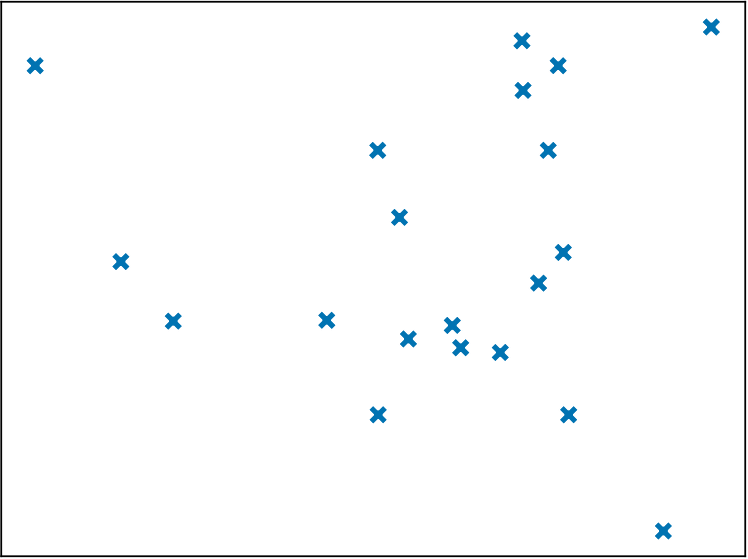}
    \end{subfigure}
    \begin{subfigure}[b]{0.225\textwidth}
        \includegraphics[width=\textwidth]{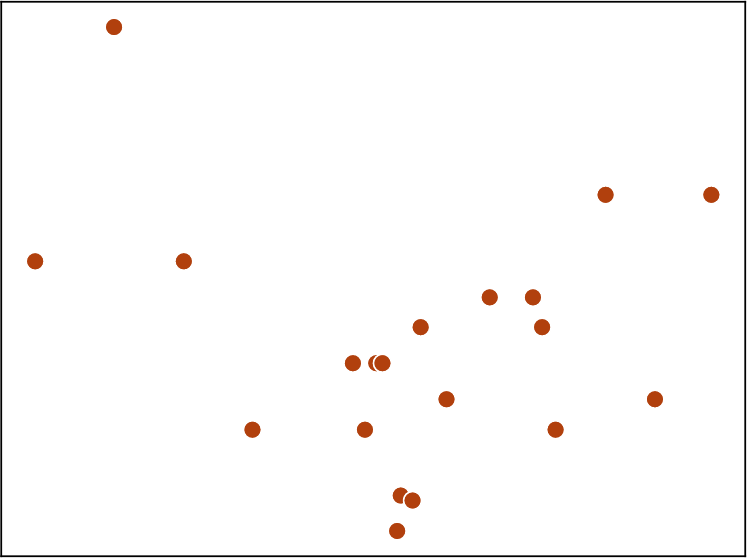}
    \end{subfigure}
    \begin{subfigure}[b]{0.225\textwidth}
        \includegraphics[width=\textwidth]{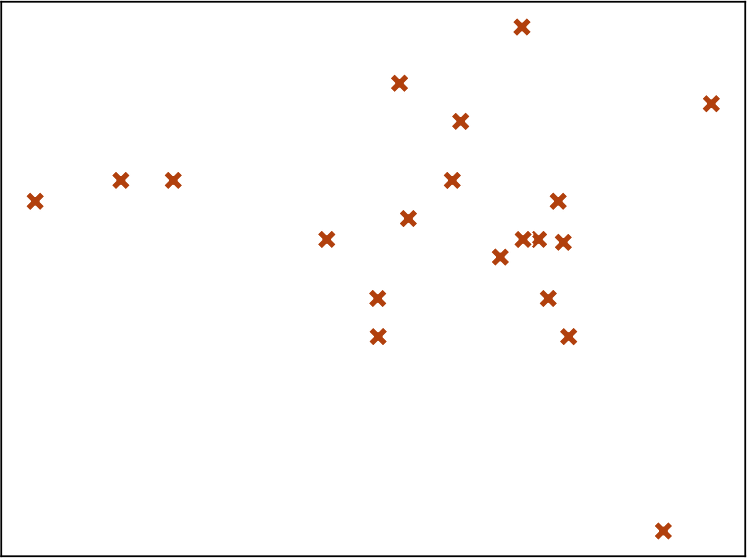}
    \end{subfigure}
    \vspace{-4.5em}
    \vspace{0.5em}
    \end{subfigure}
    \begin{subfigure}[b]{\textwidth}
     \begin{subfigure}[c]{0.07\textwidth}
     \centering
         \caption*{WEAT\\Gender\\(es)}
         \vspace{25mm}
     \end{subfigure}
    \begin{subfigure}[b]{0.225\textwidth}
        \includegraphics[width=\textwidth]{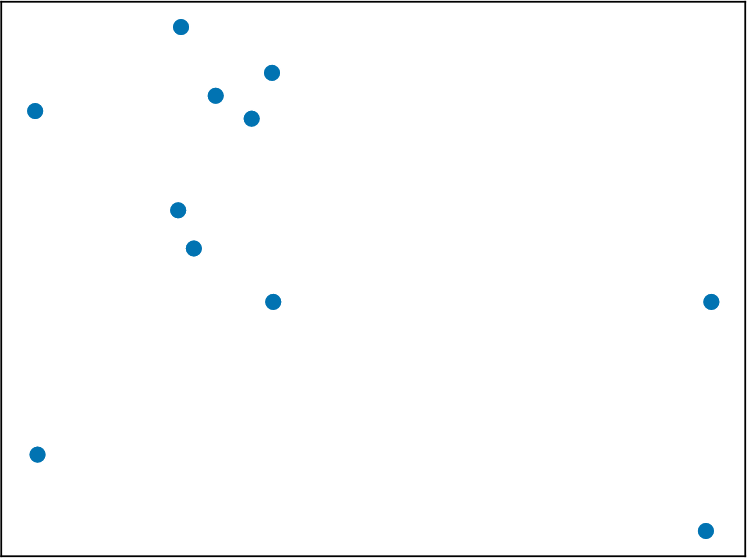}
    \end{subfigure}
    \begin{subfigure}[b]{0.225\textwidth}
        \includegraphics[width=\textwidth]{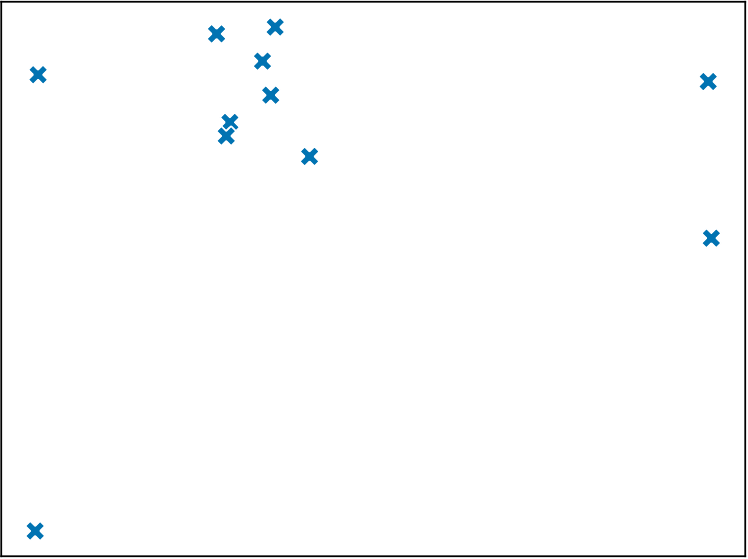}
    \end{subfigure}
    \begin{subfigure}[b]{0.225\textwidth}
        \includegraphics[width=\textwidth]{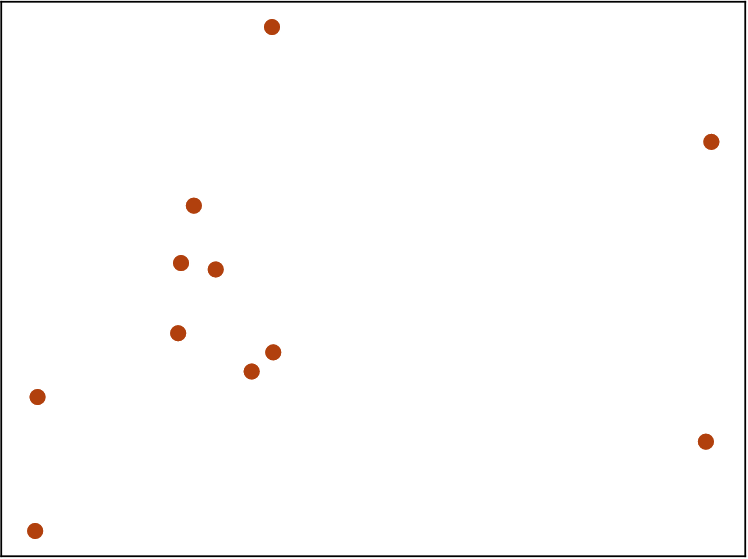}
    \end{subfigure}
    \begin{subfigure}[b]{0.225\textwidth}
        \includegraphics[width=\textwidth]{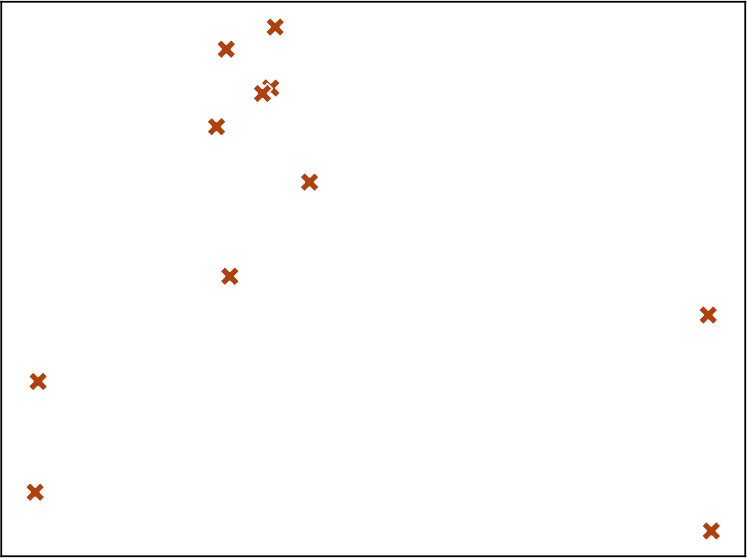}
    \end{subfigure}
    \vspace{-4em}
    \end{subfigure}
    \begin{subfigure}[b]{\textwidth}
     \begin{subfigure}[c]{0.07\textwidth}
     \centering
         \caption*{WEAT\\Migrant\\(es)}
         \vspace{25mm}
     \end{subfigure}
    \begin{subfigure}[b]{0.225\textwidth}
        \includegraphics[width=\textwidth]{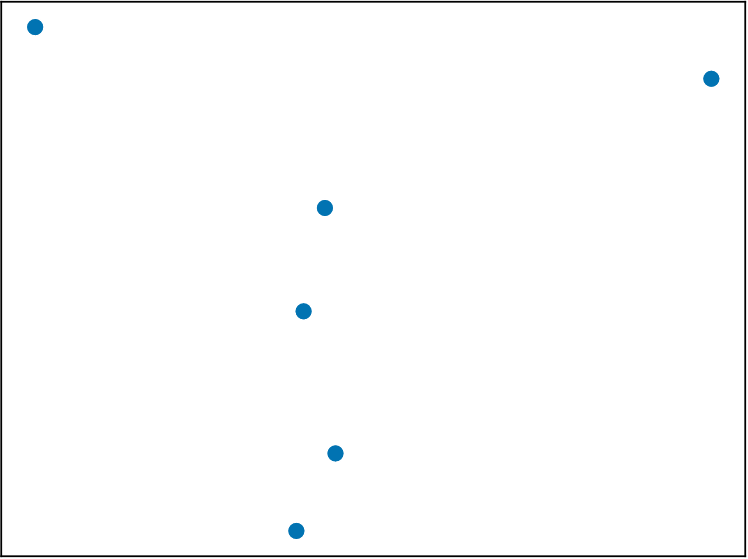}
    \end{subfigure}
    \begin{subfigure}[b]{0.225\textwidth}
        \includegraphics[width=\textwidth]{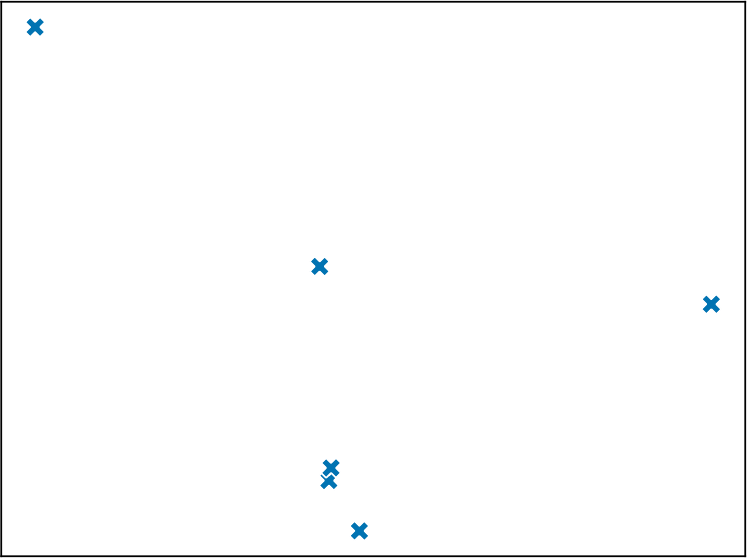}
    \end{subfigure}
    \begin{subfigure}[b]{0.225\textwidth}
        \includegraphics[width=\textwidth]{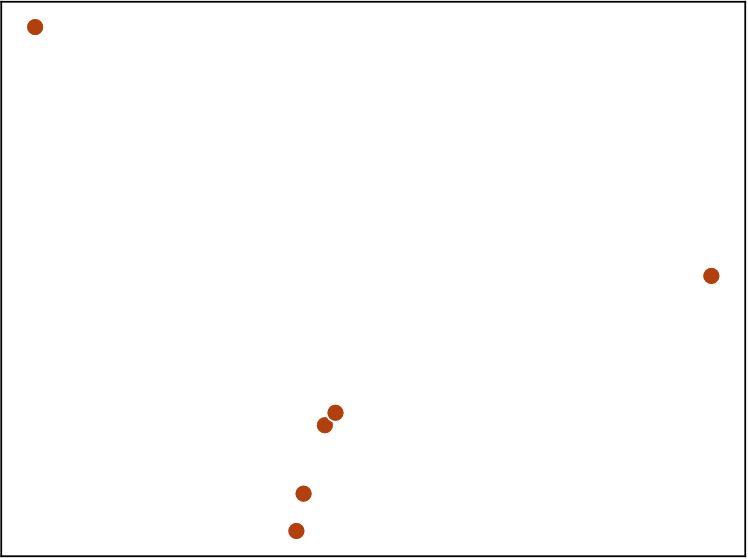}
    \end{subfigure}
    \begin{subfigure}[b]{0.225\textwidth}
        \includegraphics[width=\textwidth]{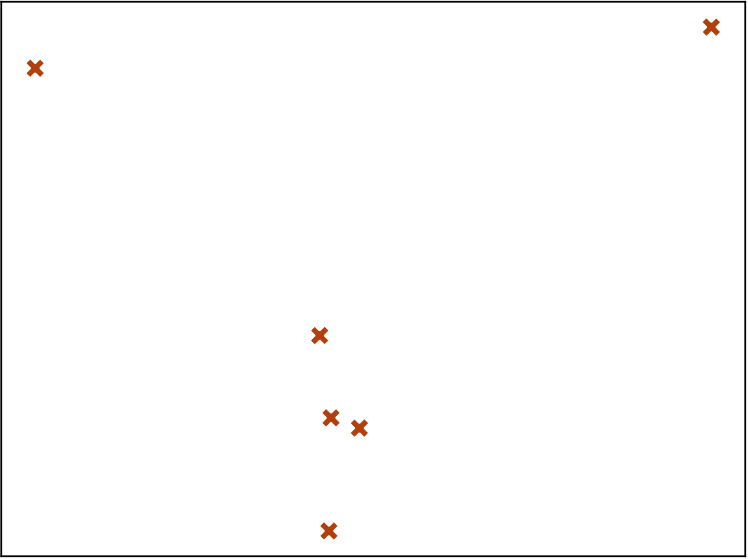}
    \end{subfigure}
    \vspace{-4em}
    \vspace{0.5em}
    \end{subfigure}

    \caption{Experimental results showing one plot per experiment, where an experiment is a triple of intrinsic metric, extrinsic metric, embedding algorithm. Tasks are English coreference (rows 1-3), English hate speech detection (rows 4-6) and Spanish hate speech detection (rows 7-8).
    }
    \label{fig:all_results}
    \vspace{-1em}
\end{figure*}
}

\begin{figure*}[t!]
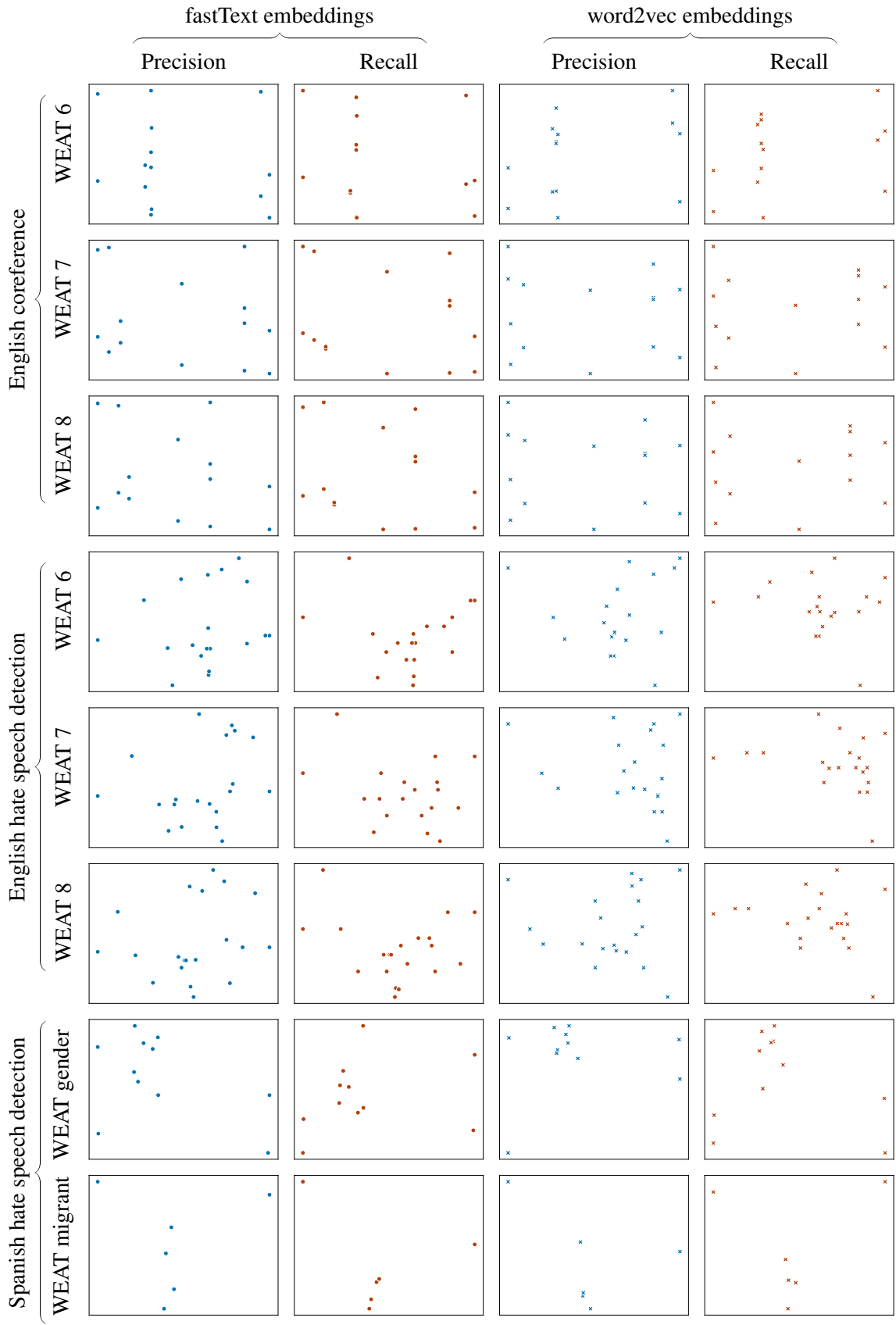

    \centering
    \begin{tikzpicture}
    \matrix (m) [matrix of nodes]{
         & Precision & Recall & Precision & Recall\\
        |[yshift=7mm]| \rotatebox{90}{WEAT 6} & \includegraphics[width=0.2\textwidth]{coref_fastText_Precision_6-cropped.pdf} & \includegraphics[width=0.2\textwidth]{coref_fastText_Recall_6-cropped.pdf} & \includegraphics[width=0.2\textwidth]{coref_word2vec_Precision_6-cropped.pdf} & \includegraphics[width=0.2\textwidth]{coref_word2vec_Recall_6-cropped.pdf} \\
        |[yshift=7mm]| \rotatebox{90}{WEAT 7} & \includegraphics[width=0.2\textwidth]{coref_fastText_Precision_7-cropped.pdf} & \includegraphics[width=0.2\textwidth]{coref_fastText_Recall_7-cropped.pdf} & \includegraphics[width=0.2\textwidth]{coref_word2vec_Precision_7-cropped.pdf} & \includegraphics[width=0.2\textwidth]{coref_word2vec_Recall_7-cropped.pdf} \\
        |[yshift=7mm]| \rotatebox{90}{WEAT 8} & \includegraphics[width=0.2\textwidth]{coref_fastText_Precision_8-cropped.pdf} & \includegraphics[width=0.2\textwidth]{coref_fastText_Recall_8-cropped.pdf} & \includegraphics[width=0.2\textwidth]{coref_word2vec_Precision_8-cropped.pdf} & \includegraphics[width=0.2\textwidth]{coref_word2vec_Recall_8-cropped.pdf} \\
        |[yshift=7mm]| \rotatebox{90}{WEAT 6} & \includegraphics[width=0.2\textwidth]{hsd_en_fastText_Precision_6-cropped.pdf} & \includegraphics[width=0.2\textwidth]{hsd_en_fastText_Recall_6-cropped.pdf} & \includegraphics[width=0.2\textwidth]{hsd_en_word2vec_Precision_6-cropped.pdf} & \includegraphics[width=0.2\textwidth]{hsd_en_word2vec_Recall_6-cropped.pdf} \\
        |[yshift=7mm]| \rotatebox{90}{WEAT 7} & \includegraphics[width=0.2\textwidth]{hsd_en_fastText_Precision_7-cropped.pdf} & \includegraphics[width=0.2\textwidth]{hsd_en_fastText_Recall_7-cropped.pdf} & \includegraphics[width=0.2\textwidth]{hsd_en_word2vec_Precision_7-cropped.pdf} & \includegraphics[width=0.2\textwidth]{hsd_en_word2vec_Recall_7-cropped.pdf} \\
        |[yshift=7mm]| \rotatebox{90}{WEAT 8} & \includegraphics[width=0.2\textwidth]{hsd_en_fastText_Precision_8-cropped.pdf} & \includegraphics[width=0.2\textwidth]{hsd_en_fastText_Recall_8-cropped.pdf} & \includegraphics[width=0.2\textwidth]{hsd_en_word2vec_Precision_8-cropped.pdf} & \includegraphics[width=0.2\textwidth]{hsd_en_word2vec_Recall_8-cropped.pdf} \\
        \rotatebox{90}{WEAT gender} & \includegraphics[width=0.2\textwidth]{hsd_es_gender_fastText_Precision-cropped.pdf} & \includegraphics[width=0.2\textwidth]{hsd_es_gender_fastText_Recall-cropped.pdf} & \includegraphics[width=0.2\textwidth]{hsd_es_gender_word2vec_Precision-cropped.pdf} & \includegraphics[width=0.2\textwidth]{hsd_es_gender_word2vec_Recall-cropped.pdf} \\
        \rotatebox{90}{WEAT migrant} & \includegraphics[width=0.2\textwidth]{hsd_es_migrant_fastText_Precision-cropped.pdf} & \includegraphics[width=0.2\textwidth]{hsd_es_migrant_fastText_Recall-cropped.pdf} & \includegraphics[width=0.2\textwidth]{hsd_es_migrant_word2vec_Precision-cropped.pdf} & \includegraphics[width=0.2\textwidth]{hsd_es_migrant_word2vec_Recall-cropped.pdf} \\
    };
    \draw [decorate,decoration={calligraphic brace,amplitude=5pt}]  ([xshift=-3mm]m-4-1.south) -- ([xshift=-3mm]m-2-1.north) node[pos=0.5,label=left:{\rotatebox{90}{English coreference}}]{};

    \draw [decorate,decoration={calligraphic brace,amplitude=5pt}]  ([xshift=-3mm]m-7-1.south) -- ([xshift=-3mm]m-5-1.north) node[pos=0.5,label=left:{\rotatebox{90}{English hate speech detection}}]{};

    \draw [decorate,decoration={calligraphic brace,amplitude=5pt}]  ([xshift=-3mm]m-9-1.south) -- ([xshift=-3mm]m-8-1.north) node[pos=0.5,label=left:{\rotatebox{90}{Spanish hate speech detection}}]{};
            
    \draw [decorate,decoration={calligraphic brace,amplitude=5pt}]  ([yshift=3mm]m-1-2.west) -- ([yshift=3mm]m-1-3.east) node[pos=0.5,label=above:{fastText embeddings}] {};

    \draw [decorate,decoration={calligraphic brace,amplitude=5pt}]  ([yshift=3mm]m-1-4.west) -- ([yshift=3mm]m-1-5.east) node[pos=0.5,label=above:{word2vec embeddings}] {};
        
\end{tikzpicture}
    \caption{Experimental results, showing one scatterplot per experiment. An experiment consists of a task (outer row label), an embedding (outer column label), an intrinsic metric (inner row label), and an extrinsic metric (inner column label). Each point in a scatterplot is the intrinsic (y-axis) and extrinsic (x-axis) measure of bias for a single run, where word embeddings for each run have been debiased or overbiased using pre- or post-processing.}
    \label{fig:all_results}
\end{figure*}

Figure \ref{fig:all_results} displays data for all tasks: one scatterplot per triple of experimental variables: an intrinsic metric, an extrinsic metric, an embedding algorithm. If we want to be able to broadly use WEAT metrics for any given bias research, these graphs should each show a clear and a positive correlation. None of them do. There are no trends in correlation between the metrics that hold in all cases regardless of experimental detail, for any of the tasks.
We have additionally examined whether there are correlations within one bias modification method (pre or postprocessing) in case a difference in the way these methods modify embeddings causes differences in trends. In most cases this breakout tells the same story. The select cases where positive (and negative) correlations are present are discussed below. Further breakout graphs and combinations are included in Appendix \ref{app:granular_results}.

\paragraph{Coreference (en): Gender} The coreference task (Figure \ref{fig:all_results}, rows 1-3) doesn't display a clear correlation in all cases, and yet it has the clearest relationship of all three tasks, with a significant moderate positive correlation for both Predictive Parity (precision) and Equality of Opportunity (recall) for word2vec (columns 3 \& 4). The overall trends are muddied by the data for fastText, which does not have a significant correlation under any conditions. Both are expected: that coreference would display the strongest trends, and that fastText would display more unpredictable or weaker trends. The Winobias coreference task is as directly matched to the WEAT tests as it is possible to be - since both use common career words to measure bias. So the relationship between the two metrics is clearest here: moving female terms closer to certain career terms most directly helps a system resolve anti-stereotypical coreference chains. However, we still only see a correlation in wod2vec, not fastText. fastText may behave less predictably because of its use of subwords; when subwords are used, word representations are more interconnected.\footnote{For example, the representation of the word \textit{childish} is by design also made up of the representations for \textit{child} and \textit{ish}, but also all unigrams, bigrams, and trigrams it contains (\textit{c}, \textit{ch}, \textit{chi}, etc).} We can debias with regard to a specific word, but that word's embedding will still be influenced by all other words that share its character ngrams. It is difficult to predict how changing the composition of a training corpus will affect all words that contain a certain ngram (e.g. \textit{ch}) in them. 
For this reason, fastText may be initially more resistant to encoding biases than word2vec, as was found in \newcite{Lauscher2019AreWC}, but may also be more complex to debias. This has implications for extending this work to contextual models, which always use some form of subword unit.

\paragraph{Hatespeech (en): Gender} 
Hatespeech (en) has fewer and more restricted correlations than coreference, as can be seen in Figure \ref{fig:all_results}, rows 4-6. These plots show no relationship at all between intrinsic and extrinsic metrics. When data is broken out by bias modification method (see Figure \ref{fig:hsd_en_results_by_method} in Appendix \ref{app:granular_results}), it becomes clear that there is a moderate \textit{positive} correlation for postprocessing for recall, and the aggregate appears this way because there is a moderate \textit{negative} correlation for preprocessing. This holds for both embedding algorithms, though both positive and negative correlations are stronger for fastText. Precision displays no correlation. Note that the absolute variance in recall is much smaller than for precision, but this is still significant for each embedding algorithm individually and for both grouped together.  

Of interest for future bias research is that the baseline level of bias (premodification, from raw twitter data) in English hatespeech differs by embedding type, but \textit{only} for precision. Initial models (with unmodified embeddings) using fastText have 10 additional points of precision for male-targeted hatespeech than for female-targeted. However initial models using word2vec have the opposite bias and have 4 fewer points of precision for male-targeted than female targeted hatespeech. For recall, the two embedding algorithms are equivalent, with 6 fewer points for male-targeted hatespeech. In fact, in the recall metric there is an early indication of unreliability of the relationship we are examining between WEAT and extrinsic bias, because there is a spread of different WEAT results that map to nearly the same difference in recall.

\paragraph{Hatespeech (es): Gender and Migrant} For hatespeech in Spanish, we examine two kinds of bias separately --- gender bias and bias against migrants, in Figure \ref{fig:all_results}, rows 7,8. Neither gender bias nor migrant bias show positive correlations in any experimental conditions. 

\textbf{Gender bias} in our models is in an \textit{absolute} sense never present, since in Spanish hatespeech targeted against women is easier to identify than against others (
with F1 in the high 80s).\footnote{This is perhaps due to examples in the training data having clearer markers such as specific anti-female slurs, but is itself an interesting question.}
But there are no overall trends when this is bias is modified to be more or less extreme, and there are no positive correlations in any conditions. There is a moderate \textit{negative} correlation for precision only when looking at fastText embeddings.

\textbf{Migrant bias} similarly has no trends save in very restricted conditions broken out by bias modification type. In contrast to the gender case, hatespeech against migrants is clearly challenging to identify, with much lower F1 in the low 60s. There is a positive correlation between migrant bias and performance gap for recall with preprocessing in fastText only. This fits the expectation that fastText may be more sensitive to preprocessing than postprocessing due to subwords, as discussed above, though in the gender bias case with negative correlation it is equally sensitive to both, so it is hard to draw conclusions. Given the smaller number of datapoints for Spanish (discussed below) this is likely just noise.
To confuse the situation further, the only trends in precision are present in word2vec, and are negative correlations.

Note that all graphs for Spanish display central clusters, because it was more difficult to get an even spread of bias measures, and because Spanish has fewer data points than English. This is for a number of reasons that compound and underscore the difficulty of expanding supposedly language-agnostic techniques beyond English, even to high resourced languages like Spanish. We have only one WEAT test for each type of bias, since we made our own that carefully balanced grammatical gender, after rectifying the issues with the existing translated versions (see Section \ref{subsec:weat_es}). Bias modification is also more difficult - the richer agreement system in Spanish means that there are more surface forms of what would be one word in English.
In addition, the language model used for nearest neighbour expansion of wordlists (see Section \ref{subsec:bias_mod_wordlists}) produces predominantly formal register words from news or scientific articles, due to a less varied makeup of its training data than the English model. This makes them less well suited to debiasing twitter data specifically, and there were no readily available models that had more casual register. For bias against migrants, there is the additional challenge that wordlists are predominantly based on proper names, which are much rarer in twitter (which tends to use @ mentions instead) than in other media. 

\section{Discussion}
The broad result of this research is that changes in WEAT do not correlate with changes in application bias, and therefore that WEAT should not be used to measure progress in debiasing algorithms. We have found that even when we maximally target bias modification of an embedding, we cannot produce a reliable change in bias direction downstream. There was no pattern or correlation between tasks, for the same task in different languages, or even in most cases within one task. And we have chosen one of the simplest possible setups, with full-word embeddings and a single type of bias at a time. Real world scenarios can easily be more complicated and involve multiple types of bias or subword embeddings. Our findings also indicate that additional complexity may muddy the relationship further. For example, fastText behaved less predictably than word2vec across experiments, suggesting that if we were to expand to larger models that are fully reliant on subwords the patterns may become even less clear.

The implication of this finding is that an NLP scientist or engineer has limited options when investigating and mitigating bias. They must a) find the specific set of wordlists, embedding algorithms, downstream tasks, and bias modification methods that are together predictive of bias for the given task, language, and model or b) implement full systems to test application bias  directly, even if their work focuses on embeddings.

While the latter may seem onerous, it may not be more so than exhaustively searching for a configuration where intrinsic bias metrics are predictive. 

This underscores the importance of making good downstream bias measures available, as either approach will require these. More datasets that are collected need to be annotated with subgroup demographic and identity information --- there are very few available. More research needs to focus on creating good challenge sets to measure application bias. Additional research on more broad usage of unsupervised methods \cite{Zhao2020LOGANLG}
would also be valuable, though those also would benefit from subgroup identity annotation to make their results more interpretable.  

It is only when more of these things are readily available that we can see the true measure of the efficacy of our debiasing efforts.

We do note a limitation of this study in that all downstream tasks are discriminative classification tasks. Bias in classification is more straightforward to measure, with well established metrics, but covers allocational harms  (performance disparity), whereas the inclusion of generative models could better cover representational harms (misleading or harmful representations/portrayals) \citep{Blodgett2020LanguageI,crawford_keynote}. Concurrent research on causal mediation analysis for bias has shown that the embedding layer in open-domain generation has the strongest effect on gender bias (as compared to other layers of the network) \citep{vig_causal}. Further work could investigate whether generation tasks have display the same or different relationship to intrinsic metrics.





\section{Conclusion}
We have examined the relationship of the intrinsic bias metric WEAT to the extrinsic bias metrics of Equality of Opportunity and Predictive Parity, for multiple tasks and languages, and determined that positive correlations between them exist only in very restricted settings. In many cases there is either negative correlation or none at all. While intrinsic metrics such as WEAT remain good descriptive metrics for computational social science, and for examining bias in human texts, we advise that the NLP community not rely on them for measuring model bias. We instead advise that they focus on careful consideration of downstream applications and the creation of datasets and challenge sets that enable measurement at this stage.



\section*{Acknowledgements}
We thank 
Andreas Grivas,
Kate McCurdy,
Yevgen Matusevych,
Elizabeth Nielsen,
Ramon Sanabria,
Ida Szubert,
Sabine Weber, Björn Ross, Agostina Calabrese, and Eddie Ungless
for comments on earlier drafts of this paper. This work received funding from the EPSRC project EP/T023767/1.

\clearpage

\bibliography{custom, anthology}

\begin{thebibliography}{43}
\expandafter\ifx\csname natexlab\endcsname\relax\def\natexlab#1{#1}\fi

\bibitem[{Basile et~al.(2019)Basile, Bosco, Fersini, Nozza, Patti, Pardo,
  Rosso, and Sanguinetti}]{Basile2019SemEval2019T5}
Valerio Basile, C.~Bosco, E.~Fersini, Debora Nozza, V.~Patti, F.~Pardo,
  P.~Rosso, and M.~Sanguinetti. 2019.
\newblock Semeval-2019 task 5: Multilingual detection of hate speech against
  immigrants and women in twitter.
\newblock In \emph{SemEval@NAACL-HLT}.

\bibitem[{Blodgett et~al.(2020)Blodgett, Barocas, Daum{\'e}~III, and
  Wallach}]{Blodgett2020LanguageI}
Su~Lin Blodgett, Solon Barocas, Hal Daum{\'e}~III, and Hanna Wallach. 2020.
\newblock \href {https://doi.org/10.18653/v1/2020.acl-main.485} {Language
  (technology) is power: A critical survey of {``}bias{''} in {NLP}}.
\newblock In \emph{Proceedings of the 58th Annual Meeting of the Association
  for Computational Linguistics}, pages 5454--5476, Online. Association for
  Computational Linguistics.

\bibitem[{Bojanowski et~al.(2017)Bojanowski, Grave, Joulin, and
  Mikolov}]{bojanowski2017enriching}
Piotr Bojanowski, Edouard Grave, Armand Joulin, and Tomas Mikolov. 2017.
\newblock Enriching word vectors with subword information.
\newblock \emph{Transactions of the Association for Computational Linguistics},
  5:135--146.

\bibitem[{Bolukbasi et~al.(2016)Bolukbasi, Chang, Zou, Saligrama, and
  Kalai}]{Bolukbasi2016ManIT}
Tolga Bolukbasi, Kai-Wei Chang, James~Y. Zou, Venkatesh Saligrama, and
  Adam~Tauman Kalai. 2016.
\newblock Man is to computer programmer as woman is to homemaker? debiasing
  word embeddings.
\newblock In \emph{NIPS}.

\bibitem[{Caliskan et~al.(2017)Caliskan, Bryson, and
  Narayanan}]{Caliskan2017SemanticsDA}
Aylin Caliskan, Joanna~J Bryson, and Arvind Narayanan. 2017.
\newblock Semantics derived automatically from language corpora contain
  human-like biases.
\newblock \emph{Science}, 356:183--186.

\bibitem[{Crawford(2017)}]{crawford_keynote}
Kate Crawford. 2017.
\newblock \href {https://www.youtube.com/watch?v=fMym_BKWQzk} {The trouble with
  bias. (keynote at neurips)}.

\bibitem[{Davidson et~al.(2019)Davidson, Bhattacharya, and
  Weber}]{Davidson2019RacialBI}
Thomas Davidson, Debasmita Bhattacharya, and Ingmar Weber. 2019.
\newblock \href {https://doi.org/10.18653/v1/W19-3504} {Racial bias in hate
  speech and abusive language detection datasets}.
\newblock In \emph{Proceedings of the Third Workshop on Abusive Language
  Online}, pages 25--35, Florence, Italy. Association for Computational
  Linguistics.

\bibitem[{Devlin et~al.(2019)Devlin, Chang, Lee, and
  Toutanova}]{devlin-etal-2019-bert}
Jacob Devlin, Ming-Wei Chang, Kenton Lee, and Kristina Toutanova. 2019.
\newblock \href {https://doi.org/10.18653/v1/N19-1423} {{BERT}: Pre-training of
  deep bidirectional transformers for language understanding}.
\newblock In \emph{Proceedings of the 2019 Conference of the North {A}merican
  Chapter of the Association for Computational Linguistics: Human Language
  Technologies, Volume 1 (Long and Short Papers)}, pages 4171--4186,
  Minneapolis, Minnesota. Association for Computational Linguistics.

\bibitem[{Dixon et~al.(2018)Dixon, Li, Sorensen, Thain, and
  Vasserman}]{Dixon2018MeasuringAM}
Lucas Dixon, John Li, Jeffrey~Scott Sorensen, Nithum Thain, and Lucy Vasserman.
  2018.
\newblock Measuring and mitigating unintended bias in text classification.
\newblock In \emph{AIES '18}.

\bibitem[{Ethayarajh et~al.(2019)Ethayarajh, Duvenaud, and
  Hirst}]{ethayarajh-etal-2019-understanding}
Kawin Ethayarajh, David Duvenaud, and Graeme Hirst. 2019.
\newblock \href {https://doi.org/10.18653/v1/P19-1166} {Understanding
  undesirable word embedding associations}.
\newblock In \emph{Proceedings of the 57th Annual Meeting of the Association
  for Computational Linguistics}, pages 1696--1705, Florence, Italy.
  Association for Computational Linguistics.

\bibitem[{Faruqui et~al.(2016)Faruqui, Tsvetkov, Rastogi, and
  Dyer}]{Faruqui2016ProblemsWE}
Manaal Faruqui, Yulia Tsvetkov, Pushpendre Rastogi, and Chris Dyer. 2016.
\newblock \href {https://doi.org/10.18653/v1/W16-2506} {Problems with
  evaluation of word embeddings using word similarity tasks}.
\newblock In \emph{Proceedings of the 1st Workshop on Evaluating Vector-Space
  Representations for {NLP}}, pages 30--35, Berlin, Germany. Association for
  Computational Linguistics.

\bibitem[{Founta et~al.(2018)Founta, Djouvas, Chatzakou, Leontiadis, Blackburn,
  Stringhini, Vakali, Sirivianos, and Kourtellis}]{Founta2018LargeSC}
Antigoni Founta, Constantinos Djouvas, Despoina Chatzakou, Ilias Leontiadis,
  Jeremy Blackburn, Gianluca Stringhini, Athena Vakali, Michael Sirivianos, and
  Nicolas Kourtellis. 2018.
\newblock \href {https://aaai.org/ocs/index.php/ICWSM/ICWSM18/paper/view/17909}
  {Large scale crowdsourcing and characterization of twitter abusive behavior}.
\newblock In \emph{ICWSM}.

\bibitem[{Gehman et~al.(2020)Gehman, Gururangan, Sap, Choi, and
  Smith}]{Gehman2020RealToxicityPromptsEN}
Samuel Gehman, Suchin Gururangan, Maarten Sap, Yejin Choi, and Noah~A. Smith.
  2020.
\newblock Realtoxicityprompts: Evaluating neural toxic degeneration in language
  models.
\newblock In \emph{EMNLP}.

\bibitem[{Glavas et~al.(2019)Glavas, Litschko, Ruder, and
  Vulic}]{Glavas2019HowT}
Goran Glavas, Robert Litschko, Sebastian Ruder, and Ivan Vulic. 2019.
\newblock How to (properly) evaluate cross-lingual word embeddings: On strong
  baselines, comparative analyses, and some misconceptions.
\newblock In \emph{ACL}.

\bibitem[{Gonen and Goldberg(2019)}]{Gonen2019LipstickOA}
Hila Gonen and Yoav Goldberg. 2019.
\newblock Lipstick on a pig: Debiasing methods cover up systematic gender
  biases in word embeddings but do not remove them.
\newblock In \emph{NAACL-HLT}.

\bibitem[{Gonen et~al.(2019)Gonen, Kementchedjhieva, and
  Goldberg}]{Gonen2019HowDG}
Hila Gonen, Yova Kementchedjhieva, and Yoav Goldberg. 2019.
\newblock \href {https://doi.org/10.18653/v1/K19-1043} {How does grammatical
  gender affect noun representations in gender-marking languages?}
\newblock In \emph{Proceedings of the 23rd Conference on Computational Natural
  Language Learning (CoNLL)}, pages 463--471, Hong Kong, China. Association for
  Computational Linguistics.

\bibitem[{Greenwald et~al.(1998)Greenwald, McGhee, and
  Schwartz}]{Greenwald1998MeasuringID}
A.~Greenwald, D.~McGhee, and J.~L. Schwartz. 1998.
\newblock Measuring individual differences in implicit cognition: the implicit
  association test.
\newblock \emph{Journal of personality and social psychology}, 74 6:1464--80.

\bibitem[{Hardt et~al.(2016)Hardt, Price, and Srebro}]{Hardt2016EqualityOO}
Moritz Hardt, Eric Price, and Nathan Srebro. 2016.
\newblock Equality of opportunity in supervised learning.
\newblock In \emph{NIPS}.

\bibitem[{Hill et~al.(2015)Hill, Reichart, and Korhonen}]{Hill2015SimLex999ES}
Felix Hill, Roi Reichart, and A.~Korhonen. 2015.
\newblock Simlex-999: Evaluating semantic models with (genuine) similarity
  estimation.
\newblock \emph{Computational Linguistics}, 41:665--695.

\bibitem[{Hutchinson and Mitchell(2019)}]{Hutchinson201950YO}
Ben Hutchinson and Margaret Mitchell. 2019.
\newblock 50 years of test (un)fairness: Lessons for machine learning.
\newblock In \emph{FAT* '19}.

\bibitem[{Kim(2014)}]{Kim2014ConvolutionalNN}
Yoon Kim. 2014.
\newblock Convolutional neural networks for sentence classification.
\newblock In \emph{EMNLP}.

\bibitem[{Kurita et~al.(2019)Kurita, Vyas, Pareek, Black, and
  Tsvetkov}]{Kurita2019QuantifyingSB}
Keita Kurita, N.~Vyas, Ayush Pareek, A.~Black, and Yulia Tsvetkov. 2019.
\newblock Quantifying social biases in contextual word representations.
\newblock In \emph{1st ACL Workshop on Gender Bias for Natural Language
  Processing}.

\bibitem[{Lauscher and Glavas(2019)}]{Lauscher2019AreWC}
Anne Lauscher and Goran Glavas. 2019.
\newblock Are we consistently biased? multidimensional analysis of biases in
  distributional word vectors.
\newblock In \emph{*SEM@NAACL-HLT}.

\bibitem[{Lauscher et~al.(2020)Lauscher, Glavas, Ponzetto, and
  Vulic}]{Lauscher2020AGF}
Anne Lauscher, Goran Glavas, Simone~Paolo Ponzetto, and Ivan Vulic. 2020.
\newblock A general framework for implicit and explicit debiasing of
  distributional word vector spaces.
\newblock In \emph{AAAI}.

\bibitem[{Lee et~al.(2017)Lee, He, Lewis, and Zettlemoyer}]{Lee2017EndtoendNC}
Kenton Lee, Luheng He, M.~Lewis, and Luke Zettlemoyer. 2017.
\newblock End-to-end neural coreference resolution.
\newblock In \emph{EMNLP}.

\bibitem[{May et~al.(2019)May, Wang, Bordia, Bowman, and
  Rudinger}]{May2019OnMS}
Chandler May, Alex Wang, Shikha Bordia, Samuel~R. Bowman, and Rachel Rudinger.
  2019.
\newblock On measuring social biases in sentence encoders.
\newblock In \emph{NAACL-HLT}.

\bibitem[{McCurdy and Serbetci(2017)}]{McCurdy2017GrammaticalGA}
K.~McCurdy and Oguz Serbetci. 2017.
\newblock Grammatical gender associations outweigh topical gender bias in
  crosslinguistic word embeddings.
\newblock \emph{WiNLP}.

\bibitem[{Mikolov et~al.(2013)Mikolov, Chen, Corrado, and
  Dean}]{Mikolov2013EfficientEO}
Tomas Mikolov, Kai Chen, Gregory~S. Corrado, and Jeffrey Dean. 2013.
\newblock Efficient estimation of word representations in vector space.
\newblock \emph{CoRR}, abs/1301.3781.

\bibitem[{Mrksic et~al.(2017)Mrksic, Vulic, S{\'e}aghdha, Leviant, Reichart,
  Gasic, Korhonen, and Young}]{Mrksic2017SemanticSO}
Nikola Mrksic, Ivan Vulic, Diarmuid~{\'O} S{\'e}aghdha, Ira Leviant, Roi
  Reichart, Milica Gasic, Anna Korhonen, and Steve~J. Young. 2017.
\newblock Semantic specialization of distributional word vector spaces using
  monolingual and cross-lingual constraints.
\newblock \emph{Transactions of the Association for Computational Linguistics},
  5:309--324.

\bibitem[{Salamanca and Pereira(2013)}]{SALAMANCA2013}
Gastã Salamanca and Lidia Pereira. 2013.
\newblock \href
  {https://scielo.conicyt.cl/scielo.php?script=sci_arttext&pid=S0718-23762013000200003&nrm=iso}
  {{PRESTIGIO Y ESTIGMATIZACI\~A“N DE 60 NOMBRES PROPIOS EN 40 SUJETOS DE
  NIVEL EDUCACIONAL SUPERIOR}}.
\newblock \emph{{Universum (Talca)}}, 28:35 -- 57.

\bibitem[{Sap et~al.(2019)Sap, Card, Gabriel, Choi, and Smith}]{Sap2019TheRO}
Maarten Sap, D.~Card, Saadia Gabriel, Yejin Choi, and Noah~A. Smith. 2019.
\newblock The risk of racial bias in hate speech detection.
\newblock In \emph{ACL}.

\bibitem[{Sedoc and Ungar(2019)}]{sedoc-ungar-2019-role}
Jo{\~a}o Sedoc and Lyle Ungar. 2019.
\newblock \href {https://doi.org/10.18653/v1/W19-3808} {The role of protected
  class word lists in bias identification of contextualized word
  representations}.
\newblock In \emph{Proceedings of the First Workshop on Gender Bias in Natural
  Language Processing}, pages 55--61, Florence, Italy. Association for
  Computational Linguistics.

\bibitem[{Sheng et~al.(2019)Sheng, Chang, Natarajan, and Peng}]{Sheng2019TheWW}
Emily Sheng, Kai-Wei Chang, Premkumar Natarajan, and Nanyun Peng. 2019.
\newblock \href {https://doi.org/10.18653/v1/D19-1339} {The woman worked as a
  babysitter: On biases in language generation}.
\newblock In \emph{Proceedings of the 2019 Conference on Empirical Methods in
  Natural Language Processing and the 9th International Joint Conference on
  Natural Language Processing (EMNLP-IJCNLP)}, pages 3407--3412, Hong Kong,
  China. Association for Computational Linguistics.

\bibitem[{Stanovsky et~al.(2019)Stanovsky, Smith, and
  Zettlemoyer}]{Stanovsky2019EvaluatingGB}
Gabriel Stanovsky, Noah~A. Smith, and Luke Zettlemoyer. 2019.
\newblock Evaluating gender bias in machine translation.
\newblock In \emph{ACL}.

\bibitem[{Tatman(2017)}]{Tatman2017GenderAD}
Rachael Tatman. 2017.
\newblock Gender and dialect bias in youtube's automatic captions.
\newblock In \emph{EthNLP@EACL}.

\bibitem[{Vig et~al.(2020)Vig, Gehrmann, Belinkov, Qian, Nevo, Singer, and
  Shieber}]{vig_causal}
Jesse Vig, Sebastian Gehrmann, Yonatan Belinkov, Sharon Qian, Daniel Nevo,
  Yaron Singer, and Stuart Shieber. 2020.
\newblock \href
  {https://proceedings.neurips.cc/paper/2020/file/92650b2e92217715fe312e6fa7b90d82-Paper.pdf}
  {Investigating gender bias in language models using causal mediation
  analysis}.
\newblock In \emph{Advances in Neural Information Processing Systems},
  volume~33, pages 12388--12401. Curran Associates, Inc.

\bibitem[{Weischedel et~al.(2017)Weischedel, Hovy, Marcus, and
  Palmer}]{Weischedel2017OntoNotesA}
R.~Weischedel, E.~Hovy, M.~Marcus, and Martha Palmer. 2017.
\newblock Ontonotes : A large training corpus for enhanced processing.

\bibitem[{Zhao and Chang(2020)}]{Zhao2020LOGANLG}
Jieyu Zhao and Kai-Wei Chang. 2020.
\newblock \href {https://doi.org/10.18653/v1/2020.emnlp-main.155} {{LOGAN}:
  Local group bias detection by clustering}.
\newblock In \emph{Proceedings of the 2020 Conference on Empirical Methods in
  Natural Language Processing (EMNLP)}, pages 1968--1977, Online. Association
  for Computational Linguistics.

\bibitem[{Zhao et~al.(2020)Zhao, Mukherjee, Hosseini, Chang, and
  Awadallah}]{Zhao2020GenderBI}
Jieyu Zhao, Subhabrata Mukherjee, Saghar Hosseini, Kai-Wei Chang, and
  Ahmed~Hassan Awadallah. 2020.
\newblock Gender bias in multilingual embeddings and cross-lingual transfer.
\newblock In \emph{ACL}.

\bibitem[{Zhao et~al.(2019)Zhao, Wang, Yatskar, Cotterell, Ordonez, and
  Chang}]{Zhao2019GenderBI}
Jieyu Zhao, Tianlu Wang, Mark Yatskar, Ryan Cotterell, Vicente Ordonez, and
  Kai-Wei Chang. 2019.
\newblock \href {https://doi.org/10.18653/v1/N19-1064} {Gender bias in
  contextualized word embeddings}.
\newblock In \emph{Proceedings of the 2019 Conference of the North {A}merican
  Chapter of the Association for Computational Linguistics: Human Language
  Technologies, Volume 1 (Long and Short Papers)}, pages 629--634, Minneapolis,
  Minnesota. Association for Computational Linguistics.

\bibitem[{Zhao et~al.(2018{\natexlab{a}})Zhao, Wang, Yatskar, Ordonez, and
  Chang}]{Zhao2018GenderBI}
Jieyu Zhao, Tianlu Wang, Mark Yatskar, Vicente Ordonez, and Kai-Wei Chang.
  2018{\natexlab{a}}.
\newblock Gender bias in coreference resolution: Evaluation and debiasing
  methods.
\newblock In \emph{NAACL-HLT}.

\bibitem[{Zhao et~al.(2018{\natexlab{b}})Zhao, Zhou, Li, Wang, and
  Chang}]{Zhao2018LearningGW}
Jieyu Zhao, Yichao Zhou, Z.~Li, W.~Wang, and Kai-Wei Chang. 2018{\natexlab{b}}.
\newblock Learning gender-neutral word embeddings.
\newblock In \emph{EMNLP}.

\bibitem[{Zhou et~al.(2019)Zhou, Shi, Zhao, Huang, Chen, Cotterell, and
  Chang}]{Zhou2019ExaminingGB}
Pei Zhou, Weijia Shi, Jieyu Zhao, Kuan-Hao Huang, Muhao Chen, Ryan Cotterell,
  and Kai-Wei Chang. 2019.
\newblock Examining gender bias in languages with grammatical gender.
\newblock In \emph{EMNLP/IJCNLP}.

\end{thebibliography}
\bibliographystyle{acl_natbib}

\clearpage

\appendix
\section{Bias Metric Definitions \& Formulas}
\label{app:bias_metrics}
Performance Gap metrics measure difference in performance across different demographic splits of the data, and are in our case (and most commonly) applied to classification tasks.\\\\
Where $A$ is a demographic variable (race, gender, etc), $Y$ is the true label, and $\hat{Y}$ is the predicted label, a fair system will satisfy:

\begin{align*}
\scalebox{0.9}{
$P(\hat{Y} = 1 | A=x,Y=1) = P(\hat{Y} = 1 | A=y,Y=1)$}
\end{align*}
where $x$ and $y$ are demographic values usually of an \textit{privileged} and a \textit{underprivileged} group. This expresses that the probability of a given test sample being correctly identified as a true positive should be equal regardless of group, and is known as \textbf{Equality of Opportunity} \cite{Hardt2016EqualityOO}.\\
A fair system will also satisfy:
\begin{align*}
\scalebox{0.9}{
$P(\hat{Y} = 1 | A=x,Y=0) = P(\hat{Y} = 1 | A=y,Y=0)$}
\end{align*}
which expresses that that probability of a given test sample being incorrectly identified as positive is equal regardless of group. This is known as \textbf{Predictive Parity} and when combined with Equality of Opportunity is known as \textbf{Equalized Odds}.\\\\
These are easily measured in most NLP systems. The former is captured by measuring recall gap, where if $x$ is the privileged group and $y$ the underprivileged, unfairness is captured by $Recall_x - Recall_y$, where any positive value is unfair. The latter is captured by $Precision_x - Precision_y$, again where positive values are unfair.

\section{WEAT Formula and Wordlists}
\label{app:weat}

\subsection{English WEAT lists}
All are tests for gender bias.
\subsubsection{Weat 6}
WEAT 6 was modified to use the general gender terms of 7,8 rather than proper names, because the co-reference task contains no names.
\paragraph{Male}
\textit{male, man, boy, brother, he, him, his, son}
\paragraph{Female}
\textit{female, woman, girl, sister, she, her, hers, daughter}
\paragraph{Career}
\textit{executive, management, professional, corporation, salary, office, business, career}
\paragraph{Family}
\textit{home, parents, children, family, cousins, marriage, wedding, relatives}\\\\
The original WEAT 6 uses the following male and female names as the gender terms:\\
Male: \textit{John, Paul, Mike, Kevin, Steve, Greg, Jeff, Bill}\\
Female: \textit{Amy, Joan, Lisa, Sarah, Diana, Kate, Ann, Donna}.

\subsubsection{Weat 7}
\paragraph{Male}
\textit{male, man, boy, brother, he, him, his, son}
\paragraph{Female}
\textit{female, woman, girl, sister, she, her, hers, daughter}
\paragraph{Math}
\textit{math, algebra, geometry, calculus, equations, computation, numbers, addition}
\paragraph{Art}
\textit{poetry, art, dance, literature, novel, symphony, drama, sculpture}

\subsubsection{Weat 8}
\paragraph{Male}
\textit{brother, father, uncle, grandfather, son, he, his, him}
\paragraph{Female}
\textit{sister, mother, aunt, grandmother, daughter, she, hers, her}
\paragraph{Science}
\textit{science, technology, physics, chemistry, Einstein, NASA, experiment, astronomy}
\paragraph{Art}
\textit{poetry, art, Shakespeare, dance, literature, novel, symphony, drama}
\subsection{Changes to English List}
We modify WEAT 6 to use the gender terms for WEAT 7/8 as the terms for 6, but otherwise leave terms as is.

WEAT 6 (career/family vs. male/female) uses proper names as gender terms, whereas the other two tests use more standard gender terms (she, her, he, him, mother, father). This is an artifact of replicating IAT, which introduces a confound in their comparability -- if the WEAT tests have different patterns of correlation, we don't know whether this is because of the difference in the way gender bias patterns for career/family vs. for arts/science or whether it patterns differently because of proper names vs. gender terms. This is exacerbated in our case where proper names are treated even more differently than usual both in twitter (where @mentions stand in for proper names) and in the Winobias metric that we use (where professions are used instead of proper names precisely because names contain gender information and the challenge set intends to be ambiguous).

\subsection{Spanish WEAT lists:}
Recall that we created these ourselves, the gender test with reference to both the original gender focused WEAT 6,7,8 of \newcite{Caliskan2017SemanticsDA} and the translation of \newcite{Lauscher2019AreWC}, significantly modified and extended to balance grammatical gender across sets of words. The migrant test was created with reference to the tests for racism that use African-American vs. European-American names paired with pleasant vs. unpleasant terms in WEAT 3, 4, 5, using the lists of European Spanish vs. migrant Spanish names identified by \newcite{SALAMANCA2013}.
\subsubsection{Gender}
\paragraph{Male:} 
\textit{masculino, hombre, niño, hermano, él, hijo, hermano, padre, papá, tío, abuelo}
\paragraph{Female:} \textit{femenino, mujer, niña, hermana, ella, hija, hermana, madre, mamá, tía, abuela}
\paragraph{Science:}
\textit{científico, físico, químico, astrónomo, tecnológico, biólogo, científica, física, química, astrónoma, tecnológica, bióloga}
\paragraph{Art:}
\textit{arquitecto, escultor, pintor, escritor, poeta, bailarín, actor, fotógrafo, arquitecta, escultora, pintora, escritora, poetisa, bailarina, actora, fotógrafa}

\subsubsection{Migrants}
\paragraph{European-Spanish names:}
\textit{Agustina, Martina, Josefa, Antonia, Sofía, Isidora, Cristóbal, Sebastián, Agustín, Alonso, Joaquín, León, Ignacio, Julieta, Matilde}
\paragraph{Migrant-Spanish names:}
\textit{Shirley, Yamileth, Sharon, Britney, Maryori, Melody, Nayareth, Yaritza, Byron, Brian, Jason, Malcon, Justin, Jeremy, Jordan, Brayan, Yeison, Yeremi, Bairon, Yastin}
\paragraph{Pleasant terms:}
\textit{caricia, libertad, salud, amor, paz, animar, amistad, cielo, lealtad, placer, diamante, gentil, honestidad, suerte, arcoiris, diploma, regalo, honor, milagro, amanecer, familia, alegría, felicidad, risa, paraíso, vacación, paz, maravilloso, maravillosa}
\paragraph{Unpleasant terms:}
\textit{abuso, choque, suciedad, asesinato, enfermedad, accidente, muerte, sufrimiento, veneno, hedor, apestar, ataque, asalto, desastre, odio, contaminación, tragedia, divorcio, cárcel, pobreza, fea, feo, cáncer, matar, vómito, bomba, maldad, podrido, podrida, agonía, terrible, horrible, guerra, repugnante}

\section{Training Data and Preprocessing}
\label{app:train_data}
This details the data for training embeddings. For data used in training the final models, see relevant papers cited in Section \ref{subsec:datasets}.
\subsection{Wikipedia}
Wikipedia data is downloaded from the latest Wikipedia article dump, tokenized with NLTK (\url{https://www.nltk.org/}), and all words appearing less than 10 times are replaced with {\tt <unk>}. The final dataset has 439,935,872 words.

\subsection{Twitter}
Twitter data is from 2019 and is downloaded from the Internet Archive \url{https://archive.org/details/twitterstream}. Retweets are removed, and data is lowercased, tokenized with NLTK TweetTokenizer, and hashtags and @mentions are replaced with {\tt <HASH>} and {\tt <MENTION>} respectively. All words appearing less than 10 times are replaced with {\tt <unk>}. English twitter data size is 3,641,306 tweets with 38,376,060 words. Spanish twitter data size is 10,683,846 tweets with 142,715,339 words.

\section{Further Results Graphs}
\label{app:granular_results}
Below are breakouts of graphs by bias modification method, as well as full graphs with metric scales and legends.

\begin{figure*}[t!]
    \centering
    
    \begin{subfigure}[b]{\textwidth}
        \includegraphics[width=\textwidth]{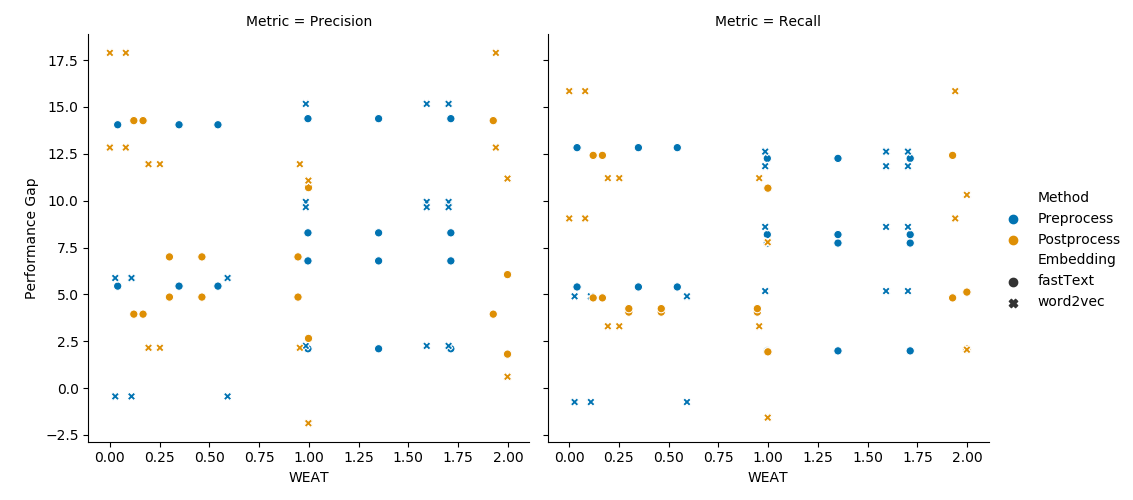}
    \caption{Coreference (en)  results broken out by bias modification method (pre- vs. post-processing).} 
    \label{fig:coref_results_by_method}
    \end{subfigure}
    
    \begin{subfigure}[b]{\textwidth}
        \includegraphics[width=\textwidth]{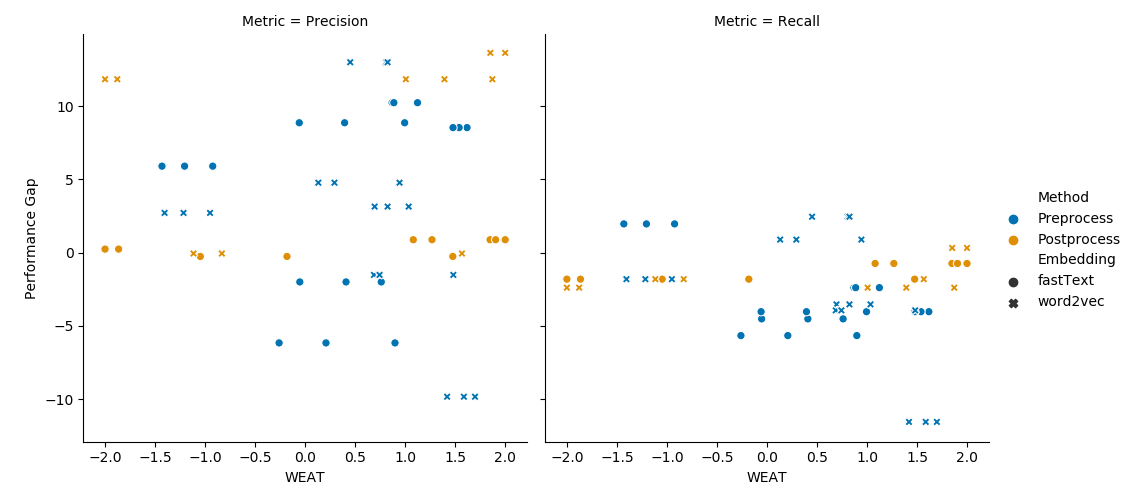}
    \caption{Hatespeech (en) results broken out by bias modification method (pre- vs. post-processing).} 
    \label{fig:hsd_en_results_by_method}
    \end{subfigure}
    
        \begin{subfigure}[b]{\textwidth}
        \includegraphics[width=\textwidth]{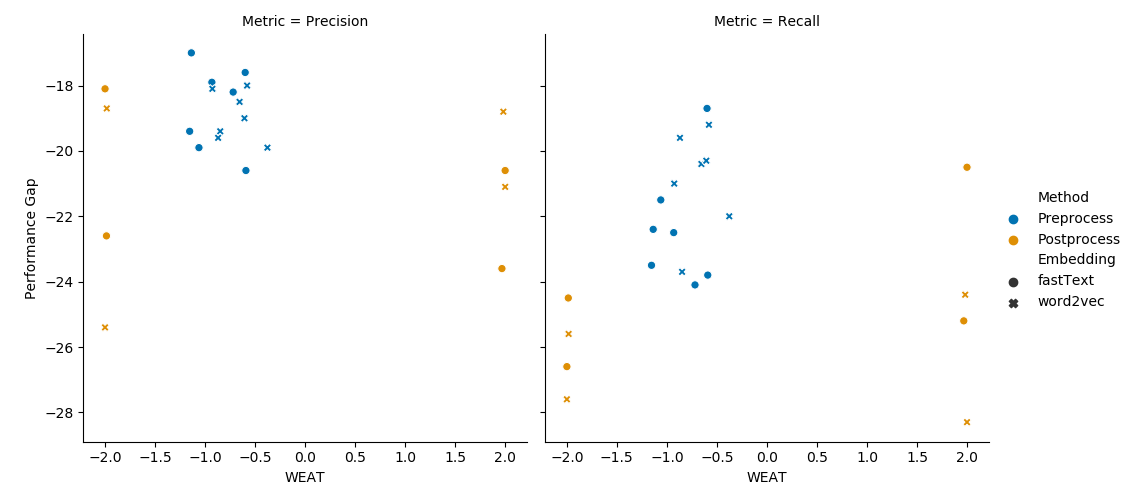}
    \caption{Hatespeech (es) results for gender bias metrics broken out by bias modification method.}    
    \label{fig:hsd_es_gender_results_by_method}
    \end{subfigure}
    
    \caption{Bias modification method breakout by pre vs. post-processing for gender bias for each task for both precision and recall.}
    \label{fig:bias_mod}
\end{figure*}

Figure \ref{fig:bias_mod} breaks out all tasks by bias modification method (pre- vs. post-processing). The main interesting thing to note here is for hatespeech in English. Based on the spread of data points, it is easy to see that there is overall more effect on precision gap when embeddings are modified, whereas recall performance gap occupies a narrower band over a wide spread of WEAT metrics.  Yet recall is the only metric which has a positive correlation with WEAT, and then only in the postprocessing condition. For Spanish it is also visible that it is much more difficult to modify bias for Spanish when preprocessing vs. when postprocessing.

\begin{figure*}[t!]
    \centering
    \begin{subfigure}[b]{0.45\textwidth}
        \includegraphics[width=\textwidth]{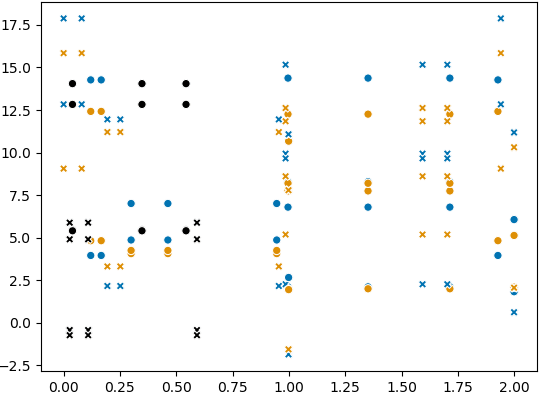}
    \caption{Coreference results (en), gender bias}
    \label{subfig:coref_app}
    \end{subfigure}
    ~ 
    \begin{subfigure}[b]{0.445\textwidth}
        \includegraphics[width=\textwidth]{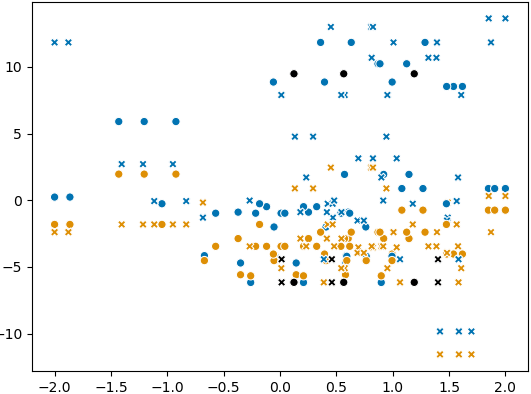}
        \caption{Hatespeech detection results (en), gender bias}
        \label{subfig:hsd_en_app}
    \end{subfigure}
    
    ~ 
    \begin{subfigure}[b]{0.45\textwidth}
        \includegraphics[width=\textwidth]{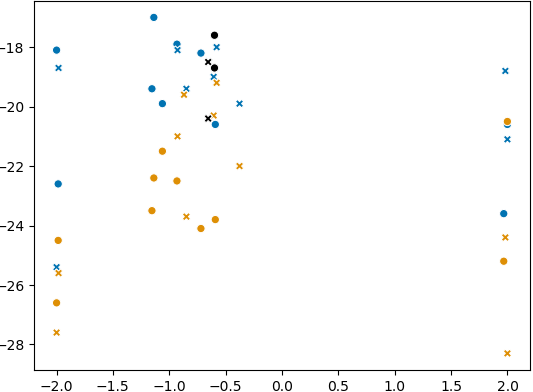}
        \caption{Hatespeech detection results (es), gender bias}
        \label{subfig:hsd_es_gen_app}
    \end{subfigure}
    ~
    \begin{subfigure}[b]{0.45\textwidth}
        \includegraphics[width=\textwidth]{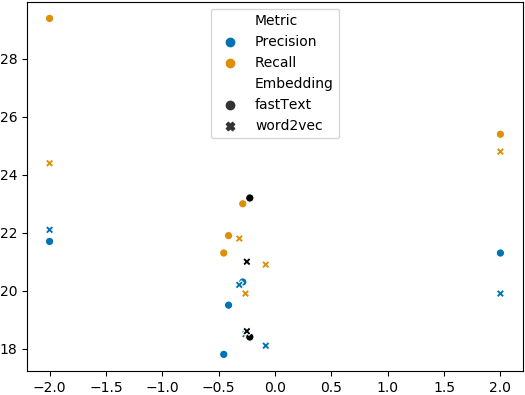}
        \caption{Hatespeech detection results (es), migrant bias}
        \label{subfig:hsd_es_mig_app}
    \end{subfigure}
    \caption{Scatterplots showing all data points for each of the 4 tasks: gender bias in co-reference (en), gender bias in hatespeech detection (en), gender bias in hatespeech detection (es), and migrant bias in hatespeech detection (es). In each plot, the $x$-axis represents WEAT, and the $y$-axis shows performance gap between groups (male-female, female-other, migrant-other). Original embeddings (before modification) shown in black. There is no correlation that holds independently of experimental conditions (embedding type, bias modification method, WEAT test).}
    \label{fig:all_results_app}
    \vspace{-1em}
\end{figure*}

Figure \ref{fig:all_results_app} shows one graph for each task and bias type combination, in full, in order to view the effect of not controlling for experimental variable. It also shows the scale for the spread of data points.

\begin{figure*}[t!]
    \centering
        \includegraphics[width=\textwidth]{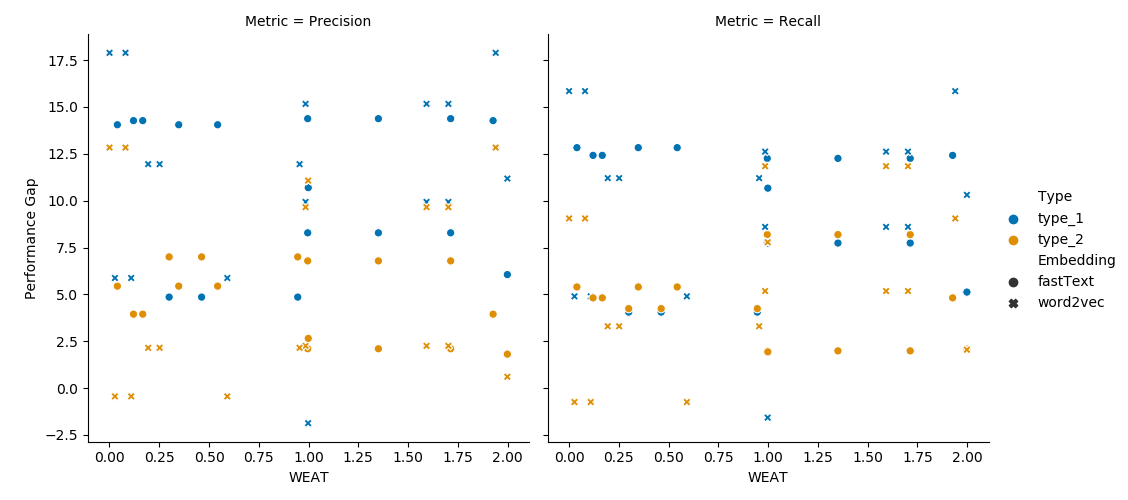}
    \caption{Coreference (en) results broken out by type of Winobias challenge, Type 1 is more difficult as there are only semantic cues to correct coreference, Type 2 has also syntactic cues.} 
  
    \label{fig:coref_results_by_type}
\end{figure*}

Finally, for interest, we also include Figure \ref{fig:coref_results_by_type}, which displays the correlation broken out by type of Winobias test (which differ in difficulty because Type 1 is semantic and Type 2 is syntactic).

\end{document}